\documentclass[lettersize,journal]{IEEEtran}
\usepackage{amsmath,amsfonts}
\usepackage{algorithmic}
\usepackage{algorithm}
\usepackage{array}
\usepackage[caption=false,font=normalsize,labelfont=sf,textfont=sf]{subfig}
\usepackage{textcomp}
\usepackage{stfloats}
\usepackage{url}
\usepackage{verbatim}
\usepackage{graphicx}
\usepackage{makecell}
\usepackage{graphicx}
\usepackage{multirow}
\usepackage[table,xcdraw]{xcolor}
\usepackage[normalem]{ulem}
\useunder{\uline}{\ul}{}
\usepackage{makecell}
\usepackage{arydshln}
\usepackage{tabularx}
\usepackage{cite}
\usepackage{bbding}
\usepackage{booktabs}
\usepackage{hyperref}
\usepackage{tikz}
\definecolor{lime}{HTML}{A6CE39}
\DeclareRobustCommand{\orcidicon}{%
    \begin{tikzpicture}
    \draw[lime, fill=lime] (0,0) 
    circle [radius=0.16] 
    node[white] {{\fontfamily{qag}\selectfont \tiny ID}};    \draw[white, fill=white] (-0.0625,0.095) 
    circle [radius=0.007];    \end{tikzpicture}
    \hspace{-2mm}}
\foreach \x in {A, ..., Z}{%
    \expandafter\xdef\csname orcid\x\endcsname{\noexpand\href{https://orcid.org/\csname orcidauthor\x\endcsname}{\noexpand\orcidicon}}
    }

\begin{document}

\title{Boosting Visual Recognition in Real-world Degradations via Unsupervised Feature Enhancement Module with Deep Channel Prior}

\author{{Zhanwen~Liu$^{\ast}$\orcidA{},~Yuhang~Li$^{\ast}$\orcidB{},~Yang~Wang$^{\dagger}$\orcidC{},~Bolin~Gao,~Yisheng~An,~Xiangmo~Zhao}

\thanks{$^{\ast}$Co-first author. $^{\dagger}$Corresponding author.}
\thanks{Zhanwen Liu, Yuhang Li, Yang Wang, Yisheng An and Xiangmo Zhao are with the School of Information Engineering, Chang'an University, Shaanxi, Xi’an 710000, China (e-mail: zwliu@chd.edu.cn; 2022124089@chd.edu.cn; ywang120@chd.edu.cn; xmzhao@chd.edu.cn).}
\thanks{Bolin Gao is with the School of Vehicle and Mobility, Tsinghua University, Beijing 100084, China (e-mail: gaobolin@tsinghua.edu.cn).}
\thanks{This work is supported by the National Natural Science Foundation of China (General Program) (No.52172302), “Two Chains” Integration Key Special Program (2023-LL-QY-24), Shannxi Province Traffic Science and Technology Program (21-02X), and the National Key R\&D Program of China (2023YFC3081700).}}

\markboth{Journal of \LaTeX\ Class Files,~Vol.~14, No.~8, August~2021}%
{Shell \MakeLowercase{\textit{et al.}}: A Sample Article Using IEEEtran.cls for IEEE Journals}


\maketitle
\begin{abstract}
The environmental perception of autonomous vehicles in normal conditions have achieved considerable success in the past decade. However, various unfavourable conditions such as fog, low-light, and motion blur will degrade image quality and pose tremendous threats to the safety of autonomous driving. That is, when applied to degraded images, state-of-the-art visual models often suffer performance decline due to the feature content loss and artifact interference caused by statistical and structural properties disruption of captured images. To address this problem, this work proposes a novel Deep Channel Prior (DCP) for degraded visual recognition. Specifically, we observe that, in the deep representation space of pre-trained models, the channel correlations of degraded features with the same degradation type have uniform distribution even if they have different content and semantics, which can facilitate the mapping relationship learning between degraded and clear representations in high-sparsity feature space. Based on this, a novel plug-and-play Unsupervised Feature Enhancement Module (UFEM) is proposed to achieve unsupervised feature correction, where the multi-adversarial mechanism is introduced in the first stage of UFEM to achieve the latent content restoration and artifact removal in high-sparsity feature space. Then, the generated features are transferred to the second stage for global correlation modulation under the guidance of DCP to obtain high-quality and recognition-friendly features. Evaluations of three tasks and eight benchmark datasets demonstrate that our proposed method can comprehensively improve the performance of pre-trained models in real degradation conditions. The source code is available at https://github.com/liyuhang166/Deep\_Channel\_Prior

\end{abstract}

\section{Introduction}
\IEEEPARstart{I}{n} recent years, significant strides have been witnessed in autonomous driving technology \cite{wang2023new}\cite{teng2023motion}\cite{teng2022hierarchical}\cite{10109207}\cite{ li2023regional}\cite{li2024intention}\cite{liu2021cascade}\cite{liu2020scale}\cite{teng2024fusionplanner}, primarily attributed to the improved reliability of vehicle environmental perception. In this context, owing to higher performance-cost ratio and installation convenience, the on-board RGB imaging camera has become one of the most widely-used sensor to guarantee the safety of vehicular navigation. However, the image acquisition process in the real world is inevitably corrupted by various factors \cite{wang2023decoupling}\cite{wang2023brightness}\cite{wang2021leveraging}, including adverse weather conditions, sensor noise, motion blur, and so on, resulting in the reduction in image visibility and the weakness in feature representation capacity of objects. Ultimately, the image quality degradation will lead to declined navigation environment perception \cite{kim2021quality}\cite{son2020urie}\cite{wang2020deep}\cite{yang2022self}\cite{liu2023enhancing}\cite{cui2023redformer}, heightening driving risks and raising safety concerns. It is thus necessary to develop available and improved visual recognition methods in real-world degradations.


\begin{figure}
    \centering
    \includegraphics[width=0.99\linewidth]{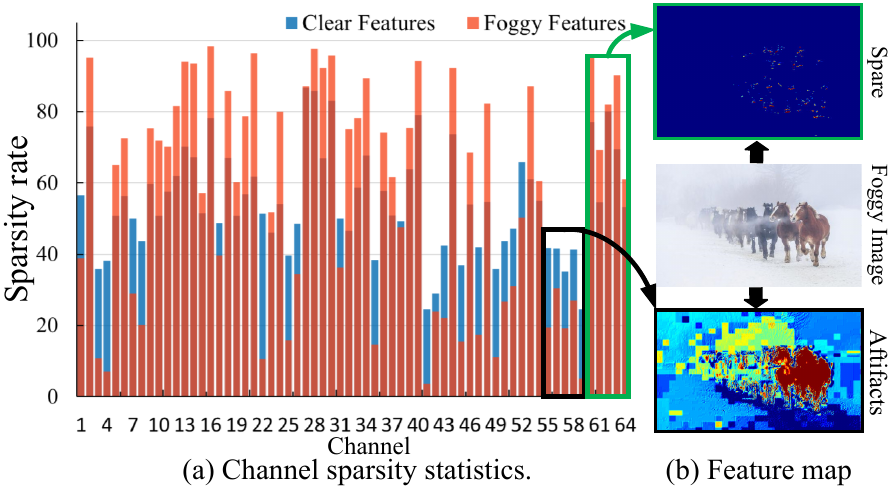}
    \caption{We randomly select 100 unpaired clear and hazy images from Haze-20 \cite{pei2018does} and extract their shallow features using the pre-trained VGG16. Then, we calculate the average sparsity (\emph{i.e.}, the proportion of pixels with zero response) of each channel and find that, degradation cues increase the overall sparsity in degraded features due to the significant loss of feature content. Besides, degradation clues will lead to the introduction of artifacts, resulting in a reduction in the sparsity of some channels.}
    \label{fig:sparsity}
\end{figure}

At present, using image restoration methods as a pre-processing to recover degraded images and then performing visual recognition is a typical solution \cite{wang2022optimal}\cite{zhao2021refinednet}\cite{zhao2022fcl}\cite{wang2017deep}. However, the existing restoration methods always prioritize visually pleasing effects and cannot guarantee consistent enhancement of structurally similar regions in images, leading to inconsistent and incomplete feature representations for recognition \cite{wang2020deep}. In response to the above challenges, recent studies have expanded the image enhancement paradigm to feature correction, aiming to directly obtain recognition-friendly features for high-level vision models \cite{kim2021quality}\cite{wang2020deep}\cite{yang2022self}. However, in the feature space, the feature maps are adaptively learned in a data-driven manner without constraints and different channels exhibit distinct spatial structures and physical meanings, presenting high spatial sparsity and channel difference. Worse, the degradation cues will result in content loss and artifacts for feature representations, which will destroy the completeness of the feature response and further aggravate spatial sparsity, as shown in Fig. \ref{fig:sparsity}. Therefore, it becomes greatly challenging to extract adequate information from the sparse and diverse features of each channel. Indeed, employing pixel-wise paired supervisory to constrain the correction process is an effective means to overcome the spatial sparsity problem \cite{kim2021quality}\cite{wang2020deep}\cite{yang2022self}. However, they struggle in real-world autonomous driving scenarios, especially when paired degraded and clear images cannot be acquired simultaneously. 

Apart from image/feature enhancement methods, Unsupervised Domain Adaptation (UDA) is also a feasible solution for feature correction, which strives to improve the model's generalization on the target domain by extracting the domain-invariant features \cite{du2021cross}\cite{long2015learning}\cite{tzeng2017adversarial}\cite{zhang2018collaborative}. However, most UDA methods require unpaired but semantic-aligned images to retrain the models. In real-world applications of autonomous driving, the integral semantic space of the degraded domain is practically inaccessible. Therefore, although UDA methods alleviate the reliance on pixel-wise supervision, limitations persist in boosting visual recognition for autonomous vehicles in the actual operating environment.


Different from existing methods, we propose a novel Deep Channel Prior (DCP) to improve the performance of visual recognition in autonomous driving within realistic degradation environments. This method is based on statistical observations that, i) in the deep representation space, the channel correlation of features with the same degradation type have a uniform and consistent distribution, even if they have different content and semantics. ii) the distributions of channel correlation have distinct margins under different degradation conditions. Deep channel prior enables us to extract compact and effective feature representations from sparse feature distributions, and facilitate the learning of mapping relationship between the degraded features and clear ones without requiring the supervision of paired data and semantic signals. Specifically, a two-stage Unsupervised Feature Enhancement Module (UFEM) is proposed, which takes degraded features as input to generate high-quality features in an unsupervised manner. In the first stage, a dual-learning architecture with multi-adversarial mechanism is introduced to recover the latent content and remove additional artifacts. The introduction of multi-adversarial mechanism effectively alleviates the discrimination challenges posed by highly sparse features and prevents the accumulation of errors in feature forward propagation. In the second stage, we convert high-dimensional and sparse features into a low-dimensional and compact relationship matrix by calculating the correlation between feature channels, and using this matrix to fine-tune the generated content in the first stage. After training as above, the obtained UFEM can work as a plug-and-play module to boost the performance of pre-trained models in real-world degradation conditions. 

The contribution can be summarized as follows:

(1) We find that in the deep representation space, content loss and artifact introduction lead to an overall mixture of sparse features with various degradation types. However, the correlations between feature channels exhibit distinct, well-bounded distributions, allowing effective perception of degradation-related properties. Accordingly, a novel deep channel prior is proposed for degraded visual recognition. 

(2) We propose an unsupervised feature enhancement module (UFEM) to sequentially perform latent content restoration and global correlation modulation under the guidance of DCP. After training on a limited set of unpaired images, the UFEM can seamlessly insert into existing models to boost their performance in real-world degradations. Ultimately, it meets the requirements for high-reliability visual perception in autonomous vehicles under extreme degradation environments.

(3) Extensive evaluations on eight benchmark datasets for degraded image classification, detection, and segmentation demonstrate that our method can achieve comprehensive improvement in real degradation conditions.
\section{Related Work}
\begin{figure*}
    \centering
    \includegraphics[width=0.96\linewidth,height=0.23\textheight]{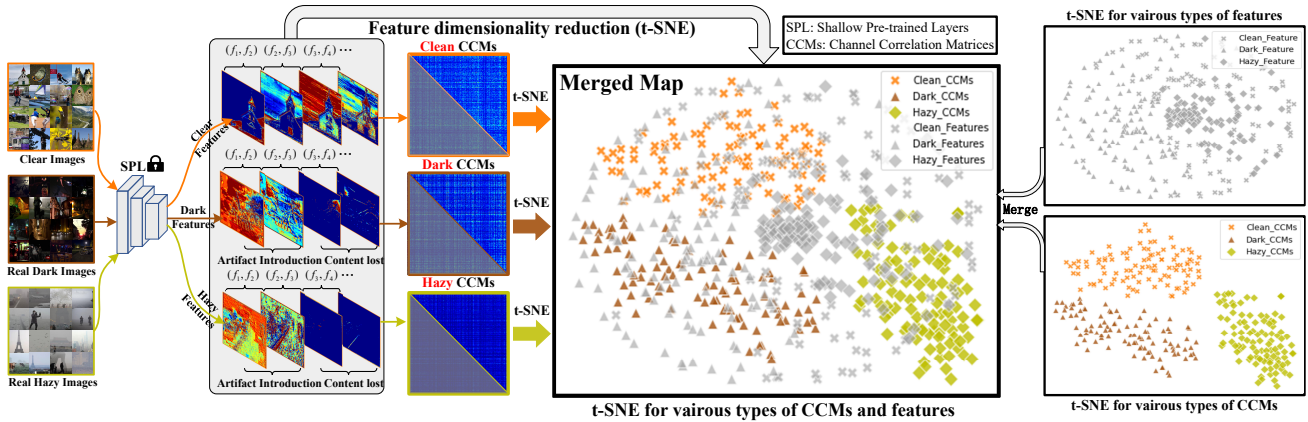}
    \caption{The illustration of Deep Channel Prior (DCP). This illustrates that the channel correlation matrix of features is an explicit means to reflect the corruption type of degraded images, while the feature itself can not represent its degradation type. The clear, dark, and hazy images are respectively from ImageNet, ExDARK, and Haze-20. And $f$ represents the feature map of a specific channel. Similar phenomena can be observed in both VGG16 and ResNet50.}
    \label{fig:gram}
\end{figure*}
\label{sec:relatedwork} 
With the increasing popularity of autonomous vehicles, the safety concerns of autonomous driving are also rising. One of the vital challenges in ensuring the safety of autonomous driving lies in improving the limited performance of its visual recognition system under various degradation scenarios. Fragility of deep neural networks for visual recognition has been studied in recent years \cite{pei2018does}\cite{loh2019getting}\cite{hendrycks2019benchmarking}\cite{10384822}, which revealed that existing recognition networks trained on HQ images are vulnerable to image distortions such as weather, noise, blur, and low-light. To address this issue, there are currently three mainstream methods as follows:


\subsection{Image Restoration for Recognition}
In real degradation environments where paired data is not available, Unsupervised Image Restoration (UIR) is an ideal choice to restore the degraded images to a clear version \cite{wang2022optimal}\cite{jiang2021enlightengan}\cite{zhao2022crnet}\cite{10313048}\cite{10375786}\cite{10409541}, such that the appearance and structure details can be easily identified by human vision. In recent years, UIR methods have been widely used to improve the robustness of visual recognition of autonomous driving in degraded environments \cite{son2020urie}\cite{guo2020zero}\cite{jiang2022unsupervised}\cite{li2021learning}. For example, Guo \emph{et al}. \cite{guo2020zero}\cite{li2021learning} formulates light enhancement as an image-specific curve estimation task and discusses the potential benefits of Zero-DCE to object detection in the dark. Jiang \emph{et al}. \cite{jiang2022unsupervised} proposes an unsupervised decomposition and correction network for low-light enhancement and assesses its effect on object detection. Apart from the domain-specific methods, a universal dual-learning architecture with the cycle-consistent loss \cite{zhu2017unpaired} has been applied to several low-level vision tasks to achieve higher-quality imaging results for autonomous vehicles, e.g. dehazing \cite{li2021you}, denoising \cite{neshatavar2022cvf}, deblurring \cite{lu2019uid}, and low-light enhancement \cite{fu2022gan}. Despite achieving exciting perceptual effects, these restoration techniques can’t always promise an improved performance for downstream high-level vision tasks. That is to say, UIR methods can't always bring the desired accuracy improvement, and may even pose a further threat to the safety of autonomous driving.

\subsection{Feature Correction for Recognition}
Recent studies \cite{kim2021quality}\cite{wang2020deep}\cite{yang2022self} have revealed that directly bridging the gap between degraded features and clear features is an effective way to improve the robustness of pre-trained models. For example, Wang \emph{et al.} proposes a Feature De-drifting Module (FDM) \cite{wang2020deep} to rectify the drifted feature responses in the shallow layers of networks, which essentially transforms the task of degraded image restoration into a feature-based reconstruction. Along this line, QualNet \cite{kim2021quality} proposes to produce HQ-like features from low-quality images via an invertible neural network, and Yang \emph{et al}. \cite{yang2022self} further estimates the uncertainty of the feature distribution, making the module learn HQ-like features better. While these methods achieve impressive results, pixel-wise paired clear features are often employed to alleviate challenges posed by sparsity in features, which are exceedingly difficult to be satisfied in real-world autonomous driving scenarios.

\subsection{Unsupervised Domain Adaptation}
UDA methods serve as another way of feature correction, aiming to transfer models across different domains by learning domain-invariant feature representations. It can be roughly divided into three categories:  (i) metric learning methods \cite{long2015learning}\cite{saito2018maximum}, which aim to minimize the divergence metrics to alleviate the domain drift, such as, maximum mean discrepancy (MMD); (ii) pseudo label based methods \cite{zhang2018collaborative}\cite{caron2018deep}, which assign pseudo-labels to target domain images using clustering methods and progressively incorporate high-confidence results into the model's supervised training; (iii) adversarial training methods \cite{du2021cross}\cite{han2022learning}, which frequently employ domain classifiers to extract domain-invariant features for robust recognition. Although UDA methods have demonstrated effective transfer across different high-quality datasets, they always require the unpaired but semantic-aligned clear set for model retraining. That is, UDA methods necessitate the prior knowledge of integral semantic space in degraded domain to construct the above clear set. Therefore, the inaccessible semantic information of degraded domain hinders the application in real-world autonomous driving environments.

To overcome the limitations of the mentioned methods in real-world autonomous driving scenarios, we introduce an unsupervised feature enhancement module (UFEM) guided by deep channel prior (DCP). The module is devised to boost the performance of existing visual recognition models in autonomous vehicles across diverse degradation environments, offering a plug-and-play solution.


\section{Proposed Method}

\subsection{Deep Channel Prior}
\label{sec:dcp}
In the feature space, we can observe that different channels exhibit distinct spatial structures and physical meanings, leading to high spatial sparsity. The introduced degradation cues will cause the loss of feature content and artifacts, further increasing the spatial sparsity of features, as shown in Fig. \ref{fig:sparsity}. High spatial sparsity makes it challenging to extract information that can describe the characteristics of degraded cues for feature correction. To illustrate the above claims, we conducted a statistical experiment on the unpaired real clear, dark, and hazy images, sourced from the ImageNet \cite{li2018benchmarking}, ExDARK \cite{loh2019getting}, and Haze-20 \cite{pei2018does} datasets, respectively, as shown in the left column of Fig. \ref{fig:gram}. Firstly, we randomly selected 100 images from each of the three real datasets. Images from different datasets exhibited no content or degradation type correlation, while those from the same dataset shared a common degradation type. Then, to obtain their feature representations in the deep representation space, we used the shallow Pre-trained Layers (SPL) of VGG16 as the feature extractor. Specifically, the features were extracted from the “Conv3\_2” layer. In fact, the “layer1” of ResNet50 can serve as an alternative. After that, we leveraged t-SNE \cite{van2008visualizing} for direct distribution visualization of the three sets of features, shown in the gray dot area of Fig. \ref{fig:gram}. As we can see, the feature distributions from haze and low-light scenarios indeed exhibit drift compared to the clear ones. However, all three feature sets are widely dispersed and mixed together, resulting in a sparse and indistinguishable cluster. This indicates that directly extracting degradation-related characteristics from these sparse features is highly challenging. 
\begin{figure*}
    \centering
    \includegraphics[width=0.9\linewidth,height=0.4\textheight]{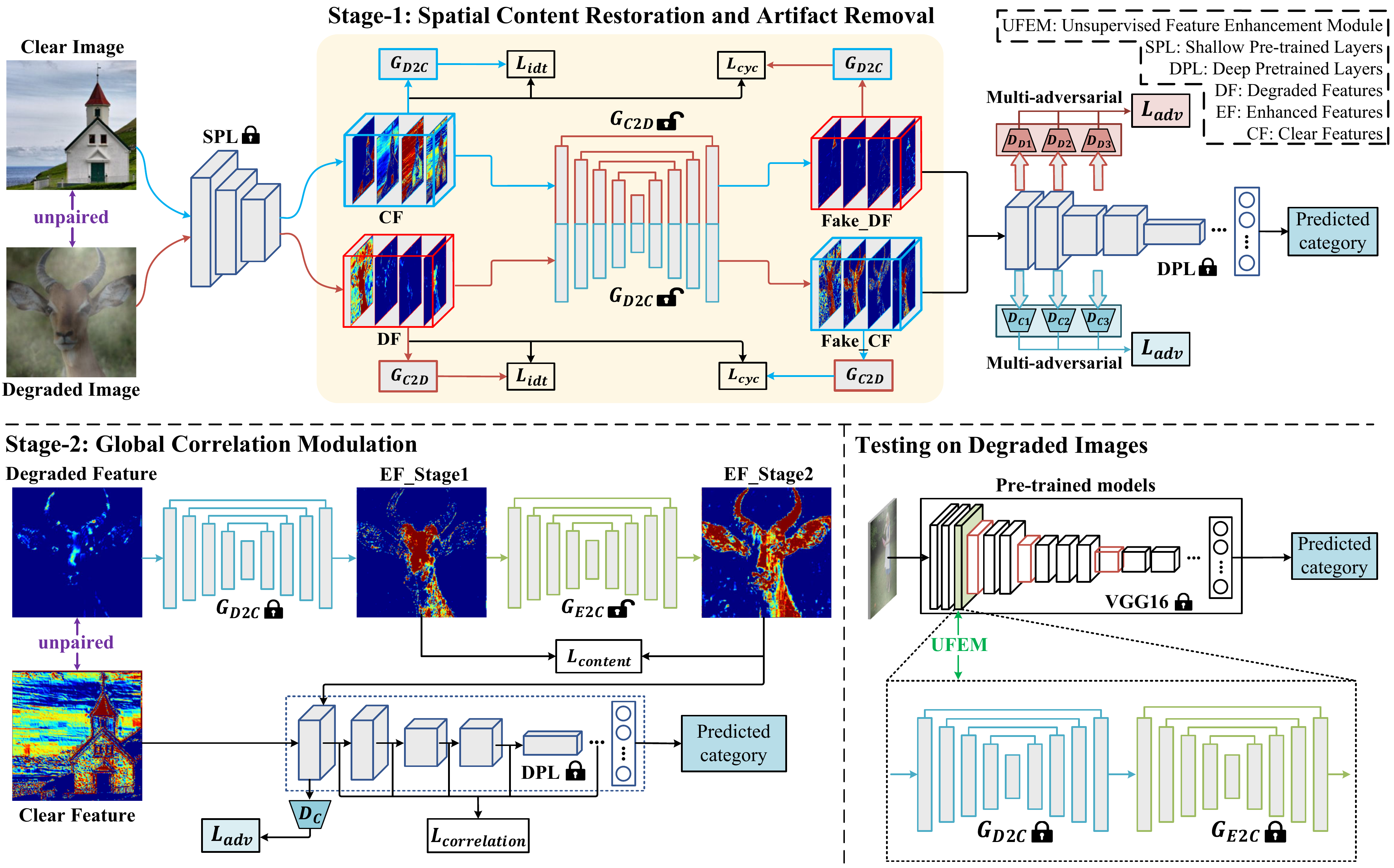}
    \caption{The pipeline of our proposed Unsupervised Feature Enhancement Module (UFEM). The UFEM takes unpaired clear features and degraded features as input, and sequentially performs content restoration and channel correlation modulation for feature correction guided by DCP. Finally, the UFEM is seamlessly inserted into existing models to improve their performance on degraded images.}
    \label{fig:framework}
\end{figure*}

To tackle this challenge, we propose to compute pairwise channel correlations for each feature set using Eq. \ref{form:gram}, resulting in three sets of channel correlation matrices, presented in the third column of Fig. \ref{fig:gram}. Notably, the channel correlation matrix $G^l \in \mathcal{R}^{N_l \times N_l}$ represents the intrinsic statistical properties of features, where $G_{i j}^l$ is the inner product between feature map $i$ and $j$ in layer $l$. Actually,  we only calculated the correlation between two channels in the feature without repetitions, and got the upper triangular part of the matrix for the final t-SNE.


\begin{equation}
\label{form:gram}
G^{l}= FF^{T}=\left[\begin{array}{ccc}
\left(f_{1}, f_{1}\right) & \cdots & \left(f_{1}, f_{l}\right) \\
\vdots & \ddots & \vdots \\
\left(f_{l}, f_{1}\right) & \cdots & \left(f_{l}, f_{l}\right)
\end{array}\right],
\end{equation}

It can be observed that: 1) channel correlation matrices from a specific type of images are clustered together, regardless of what content and semantics are contained in the images; 2) there is an obvious gap between the clear channel correlation matrices and the ones of any other degradation types. Therefore, by measuring the consistency in channel correlation matrices of features, a mapping relationship that provides an explicit optimization direction for bridging degraded features and clear ones together can be learned. In this paper, we call such a mapping as the Deep Channel Prior (DCP), which acts as a beacon in the unsupervised solution space, guiding us to explore optimal clear-like features. Next, we will present how to effectively utilize the DCP for the unsupervised correction of degraded features. 

\subsection{Content Restoration and Artifact Removal}
\label{section:stage1}
As illustrated in Fig. \ref{fig:sparsity}, clear features exhibit responsiveness in less than half of the overall area. And the interference of degradation cues results in further content loss and introduces additional artifacts in degraded features, causing higher sparsity and lower effective information ratio, exacerbating the challenge of unsupervised feature correction. To tackle this, in this paper, we propose a plug-and-play Unsupervised Feature Enhancement Module (UFEM), aiming to rectify degraded features into clear ones from both spatial and channel aspects. Firstly, in the first stage, we focus on restoring the latent content and removing additional artifacts by leveraging a dual-learning architecture. Moreover, to alleviate the discriminative challenges posed by high-dimensional sparse features, we also introduce a multi-adversarial mechanism, discriminating features from scale-smaller, information-richer, and more compact features, reducing the error accumulation in forward propagation. Specific implementation details are as follows.


\textbf{Dual-learning architecture.} We innovatively apply the dual-learning architecture \cite{zhu2017unpaired} to the shallow feature space for unsupervised feature correction, aiming to restore the latent content and remove redundant artifacts without using paired images and semantics for supervision. Taking the dehazing branch as an example, we first introduce a generator of $G_{D2C}$, striving to restore degraded features (DF, termed as $X_{D}$) to enhanced features (EF, termed as $X_{E}$), \emph{i.e.}, clear-like features. At this point, a discriminator $D_{C}$ is employed, learning to classify clear features (CF, termed as $X_{C}$) as 1 and degraded ones as 0. Following this, another generator $G_{C2D}$ is introduced to transform clear features into degraded ones. Accordingly, a discriminator $D_{D}$ is employed to classify degraded features as 1 and clear ones as 0. These above can be expressed as the adversarial loss $L_{adv}$. Besides, to ensure meaningful pairing of input degraded features and their corresponding enhanced ones, the cycle-consistent loss $L_{cyc}$ is applied, constraining the one-cycle translated results $G_{C2D}$($G_{D2C}$(DF)) to match the input DF. Finally, the identity-preserving loss $L_{idt}$ contributes to solution stability and better convergence. Mathematically, they are expressed as follows:

\begin{equation}
\label{form:adv}
L_{adv}=\log(D_C(X_C))+\log(1-D_C(G_{D2C}(X_D)),
\end{equation}
\begin{equation}
\label{form:cyc}
L_{cyc}=\|X_D-G_{C2D}(G_{D2C}(X_D))\|_1,
\end{equation}
\begin{equation}
\label{form:idt}
L_{idt}=\|X_D-G_{C2D}(X_D)\|_1,
\end{equation}

\textbf{Multi-adversarial mechanism.} As mentioned above, content loss and artifacts introduction inherently lead to the higher sparsity and less effective information in features. Hence, accurately distinguishing clear and degraded features from highly sparse features becomes challenging. To tackle this challenge, we have also incorporated a multi-adversarial mechanism into the above architecture. This addition serves to effectively address the discrimination challenges stemming from high sparsity and prevent the accumulation of errors in the propagation of enhanced features within deep networks. Specifically, as the network depth increases, the size of feature maps decreases, concentrating more meaningful information within them \cite{wang2018high}. Based on this, we go beyond the limitation of distinguishing between clear and degraded features solely in the shallow layers of translation. Instead, we introduce multi-scale discriminators in subsequent successive layers. As illustrated in Fig. \ref{fig:framework}, three independent discriminators are employed to evaluate the authenticity of enhanced features at different layers of Deep Pre-trained Layers (DPL), offering feedback to the generator. This facilitates a clearer distinction between degraded and clear features, and effectively addresses challenges arising from high sparsity and a low effective information ratio in large-sized features.

Besides, equipping discriminators with various scales constrains the propagation of enhanced features across diverse hidden layers of DPL, ensuring a more stable propagation and improved recognition for deep networks. However, layers equipped with discriminators should not be too deep to prevent the semantics in features influence on correct discrimination. We empirically choose the subsequent two layers after enhancement as the equipment layers. In the end, the adversarial loss can be extended as follows:
\begin{equation}
\label{form:adv_2}
L_{mul_{-} adv}=w_1 \cdot L_{adv_{-} 1}+w_2 \cdot L_{adv_{-} 2}+w_3 \cdot L_{adv_{-} 3},
\end{equation}

where $L_{a d v_{-} i}(i=1,2,3)$ represents the adversarial loss of different layers, and $w$ is the corresponding weight. Ultimately, our full objective function for Stage-1 is summarized as follows:
\begin{equation}
\label{form:stage1}
L_{Stage1}=\lambda_1 \cdot L_{mul\_adv}+\lambda_2 \cdot L_{cyc}+\lambda_3  \cdot L_{idt},
\end{equation}

where $\lambda$ represents the weighting coefficient. After training as above, the $G_{D2C}$ is obtained and preserved, then inserted into existing models to restore latent content and remove additional artifacts in degraded features, while the other modules are all discarded.

\subsection{Global Correlation Modulation}
\label{section:stage2}
Despite the rough restoration of latent content in degraded features, performing feature correction in a high-dimensional sparse space is still constrained by poor effective information extraction, hindering the perception of statistical properties associated with degradation cues. To tackle this, we propose a deep channel prior, which finds that in the deep representation space, the channel correlation matrices $G_{n \times n}$ ($n$ denotes the count of feature channels) of features can accurately reflect their degradation types, even if they have distinct content and semantics. Therefore, motivated by DCP, we transform the high-dimensional sparse features into low-dimensional compact matrix representations. By
learning the mapping relationship between the correlation matrices of degraded and clear features, we further refine the features with roughly restored content, thus benefiting deep network recognition.

\textbf{Implementation.} As mentioned above, we leverage the correlations between different filter responses to effectively characterize degradation types, rather than high-dimensional sparse features. These feature correlations are given by the channel correlation matrix, which converts the highly sparse features into a compact matrix representation. This facilitates a better perception of statistical properties related to degradation cues, easing the learning of the mapping relationship between degraded and clear features. Thus, by minimizing the absolute distance of the channel correlation matrices between EF (generated in Stage-1) and CF at multiple layers, we learn a mapping that can bridge the gap between enhanced and clear features. Based on this, we obtained the enhanced features $\mathrm{\widetilde{EF}}$ (generated in Stage-2) that appear as clear as possible. This can be expressed as the correlation-consistent loss:
\begin{equation}
\label{form:correlation}
L_{correlation}=\sum_{l=0}^{L} w_{l} \cdot \sum_{i, j}\left\|G_{i j}^{l}-\hat{G}_{i j}^{l}\right\|_{1},
\end{equation}

where $G^l$ and $\hat{G}^l$ are the respective channel correlation matrix representations of CF and $\mathrm{\widetilde{EF}}$ in layer $l$, and $w$ is the weight corresponding to different layers. Here, we choose the “Conv1\_2”, “Conv2\_2”, “Conv3\_3”, and “Conv4\_3” from VGG16 as examples of $L$. It is similar for other backbones.

Moreover, GANs \cite{goodfellow2014generative} are also employed to further reduce the disparity between EF and CF. In this, the generator $G_{E2C}$ endeavors to deceive the discriminator $D_{C}$ by generating more realistic clear-like features $\mathrm{\widetilde{EF}}$, while $D_{C}$ learns to classify the CF to class 1 and the $\mathrm{\widetilde{EF}}$ to class 0. This can be expressed as the adversarial loss:
\begin{equation}
\label{form:adv_3}
L_{adv}=\log(D_C(X_C))+\log(1-D_C(G_{E2C}(X_E)),
\end{equation}

Besides, since the above process of converging to a consistent distribution is unconstrained, the intrinsic semantic information restored in the EF can be easily changed. Therefore, it is necessary to ensure approximately maximal preservation of the content in EF during correlation modulation. In this paper, we enforce a constraint on the content consistency between the $\mathrm{\widetilde{EF}}$ (generated in Stage-2) and the EF (generated in Stage-1) for semantic fidelity, thereby making $\mathrm{\widetilde{EF}}$ progressively approaching a clear style while preserving content information. This can be expressed as the content-consistent loss:
\begin{equation}
\label{form:content}
L_{content}=\sum_{i, j}\left\|C_{i j}^{l}-\hat{C}_{i j}^{l}\right\|_{1},
\end{equation}

where $C^l$ and $\hat{C}^{l}$ are the respective content representations of CF and $\mathrm{\widetilde{EF}}$ in layer $l$, \emph{e.g.}, the “conv4\_3” of VGG16. In the end, our full objective function for Stage-2 is summarized as follows:
\begin{equation}
\label{form:stage2}
L_{Stage2}=\lambda_{1} \cdot L_{correlation}+\lambda_{2} \cdot L_{adv}+\lambda_{3} \cdot L_{content},
\end{equation}

where $\lambda$ is the corresponding weight of different losses. After training as above, only $G_{D2C}$ and $G_{E2C}$ are retained and frozen, comprising the UFEM, which brings approximately maximal restoration of latent content whilst achieving fine modulation of channel correlations. Finally, UFEM is seamlessly inserted into existing models, improving their performance in real-world degradations.

\section{Experiments}
\label{section:exps}

\subsection{Dataset and Implementation}
\textbf{Dataset Details.} We conduct extensive experiments for evaluating our UFEM on three kinds of high-level vision tasks in autonomous driving, \emph{i.e.}, image classification (IC), object detection (OD), and semantic segmentation (SS). Specifically, we perform evaluations on one large synthetic dataset and seven real-world degradation datasets. Tab. \ref{tab:datasets} provides a summary of crucial statistics for the all employed datasets.

\textbf{Implementation Details.} We employ pre-trained VGG16 \cite{simonyan2014very}, ResNet50 \cite{he2016deep}, YOLOv5 \cite{redmon2016you}, and DeepLabv3+ \cite{chen2018encoder} as the baseline models. The UFEM is trained for 200 epochs using the Adam optimizer with a batch size of 5. The initial learning rate of generators and discriminators are set to 2e-4 and 1e-4, respectively. The learning rate is multiplied by 0.1 every 10 epochs. All experiments are run on one Nvidia A100 GPU. More implementation details on the generator and discriminator structures can be found on our GitHub. 

\subsection{Model Fine-tuning and Method Comparison Details}
\textbf{Model Fine-tuning on Clear Dataset.} As mentioned above, we conduct comparisons on both synthetic and real datasets. Before conducting comparisons on real datasets, since the categories of Haze-20 \cite{pei2018does}, ExDARK \cite{loh2019getting}, RTTS \cite{li2018benchmarking}, and DAWN \cite{hassaballah2020vehicle} are inconsistent with the pre-trained models, we need to fine-tune the models to obtain the well-performing classifiers and detectors on clear images. For image classification on Haze-20, we first randomly select 100 clear images per class from HazeClear-20 (80 for training, 20 for validation, and the remainder for testing) to fine-tune the pretrained classifiers. After that, we obtain the classifiers for 20 classes, achieving accuracies of 95.9\% and 93.9\% on the test sets, respectively. For image classification on ExDARK, its fine-tuning is similar to the above, however, there is no clear set corresponding to ExDARK. Therefore, following the collection in \cite{loh2019getting}, we manually select images of the same categories as ExDARK from the ImageNet \cite{deng2009imagenet}, PASCAL VOC \cite{everingham2010pascal}, and COCO \cite{lin2014microsoft} datasets, filter out the obviously low-light images and ultimately obtain the clear dataset ExDark-clean for fine-tuning. For object detection on RTTS and DAWN, we choose CityScapes \cite{cordts2016cityscapes} as our clear set and filter its categories to be consistent with RTTS (5 classes) and DAWN (6 classes) respectively. After that, fine-tuning is performed on the clear set to obtain a detector that performs well on the clear set. Different from the above, for the ImageNet-C and three nighttime semantic segmentation datasets, whose categories are aligned with the official pre-trained models, we do not fine-tune but choose the pre-trained models as the baselines.

\begin{table}[tbp]
\small
\centering
\caption{Statistics of the datasets used to report results on three different downstream vision tasks.}
\label{tab:datasets}
\renewcommand\arraystretch{1.1} 
\scalebox{0.74}{
\begin{tabular}{l|c|c|c|c|c}
\textbf{Dataset}  & \textbf{Task}       & \textbf{Volume} & \textbf{Class} & \textbf{Type}  & \textbf{Degradation} \\ \hline
\textbf{ImageNet-C \cite{hendrycks2019benchmarking}}        & IC  & 50,000    & 1,000  & Synthetic          & Fog, Blur   \\
\textbf{Haze-20 \cite{pei2018does}}           & IC  & 4,610     & 20    & Real               & Fog         \\
\textbf{ExDARK \cite{loh2019getting}}            & IC  & 7,363     & 12    & Real               & Night       \\
\textbf{RTTS \cite{li2018benchmarking}}              & OD      & 4,322     & 5     & Real               & Fog         \\
\textbf{DAWN Fog \cite{hassaballah2020vehicle}}          & OD      & 302      & 6     & Real               & Fog         \\
\textbf{ACDC Nightime \cite{sakaridis2021acdc}}     & SS & 1,006     & 19    & Real               & Night       \\
\textbf{Dark\_Zurich \cite{sakaridis2019guided}}      & SS & 50       & 19    & Real               & Night       \\
\textbf{Nighttime Driving \cite{dai2018dark}} & SS & 50       & 19    & Real               & Night       \\ \Xhline{1.2pt}
\end{tabular}
}
\end{table}

\textbf{Training Details of Comparison Methods.} In this part, we further provide the detailed training settings of various comparison methods on various datasets. For UIR methods on synthetic dataset, we randomly select 1,200 pairs of unpaired images from each degradation level for model training. And, the validation is performed on a test set of 50,000 images from ImageNet-C. For UDA methods on synthetic dataset, which require retraining models with unpaired but semantic-aligned images, we select 10 pairs of unpaired images from each class of ImageNet and ImageNet-C, forming a dataset of 10K pairs of images for retraining. For UIR methods on real datasets, we select about 500 pairs of unpaired images from Haze-20 and ExDARK for the training of UIR methods. The weights obtained from Haze-20 are subsequently used for testing on RTTS and DAWN. However, for the night segmentation datasets, due to the lack of clear sets, we only utilize the pre-trained weights released by the UIR methods.

\begin{table*}[htbp]
\small
    \centering
    \caption{Quantitative comparison of our proposed UFEM with Image Restoration (IR) and Unsupervised Domain Adaption (UDA) methods on foggy and motion blur conditions, which contains 5 severity levels.  The Top-1 and Top-2 performances are marked in {\color[HTML]{fe0000}\textbf{bold}} and {\color[HTML]{3531ff}\textbf{bold}}, respectively. Adding shades of gray are the performances of our UFEM. These settings apply to all subsequent tables.}
    \label{tab:cls_syn}
    \resizebox{0.94\textwidth}{!}{
    \renewcommand\arraystretch{1.2}
    \begin{tabular}{lccccccccccc}
    \Xhline{1.5pt}
                                 & \multicolumn{1}{c||}{}                                
                                  & \multicolumn{5}{c||}{\textbf{Top-1 Accuracy of VGG16 (\%)}}           & \multicolumn{5}{c}{\textbf{Top-1 Accuracy of ResNet50 (\%)}}                                                                                                                                                                                                                                                     \\ 
                                  \cline{3-12} 

\multirow{-2}{*}{\textbf{Method}} & \multicolumn{1}{c||}{\multirow{-2}{*}{\textbf{Venue}}}  & \multicolumn{1}{c|}{\textbf{Level-1}}                           & \multicolumn{1}{c|}{\textbf{Level-2}}                           & \multicolumn{1}{c|}{\textbf{Level-3}}                            & \multicolumn{1}{c|}{\textbf{Level-4}}                            & \multicolumn{1}{c||}{\textbf{Level-5}}                           & \multicolumn{1}{c|}{\textbf{Level-1}}                           & \multicolumn{1}{c|}{\textbf{Level-2}}                           & \multicolumn{1}{c|}{\textbf{Level-3}}                           & \multicolumn{1}{c|}{\textbf{Level-4}}                           & \textbf{Level-5}                           \\ \hline
\multicolumn{12}{c}{\textbf{Degradation Type: Fog} (\textbf{100} unpaired images for UFEM, \textbf{1200} unpaired images for UIR methods, \textbf{10K} unpaired images for UDA methods)}      \\ \hline
Baseline                & \multicolumn{1}{c||}{\textbf{-}}                      & \multicolumn{1}{c|}{53.6}                                       & \multicolumn{1}{c|}{46.1}                                       & \multicolumn{1}{c|}{36.7}                                        & \multicolumn{1}{c|}{31.6}                                        & \multicolumn{1}{c||}{17.5}                                       & \multicolumn{1}{c|}{61.4}                                       & \multicolumn{1}{c|}{55.3}                                       & \multicolumn{1}{c|}{46.2}                                       & \multicolumn{1}{c|}{39.6}                                       & 23.7                                       \\
YOLY \cite{li2021you}                     & \multicolumn{1}{c||}{IJCV'21}                   & \multicolumn{1}{c|}{57.2}                                       & \multicolumn{1}{c|}{52.6}                                       & \multicolumn{1}{c|}{44.0}                                        & \multicolumn{1}{c|}{36.6}                                        & \multicolumn{1}{c||}{20.4}                                       & \multicolumn{1}{c|}{63.1}                                       & \multicolumn{1}{c|}{58.0}                                       & \multicolumn{1}{c|}{49.4}                                       & \multicolumn{1}{c|}{41.9}                                       & 25.4                                       \\
RefineDNet \cite{zhao2021refinednet}              & \multicolumn{1}{c||}{TIP'21}                    & \multicolumn{1}{c|}{58.3}                                       & \multicolumn{1}{c|}{54.8}                                       & \multicolumn{1}{c|}{48.4}                                        & \multicolumn{1}{c|}{42.1}                                        & \multicolumn{1}{c||}{\color[HTML]{3531ff} \textbf{27.6 (10.1↑)}}                          & \multicolumn{1}{c|}{62.2}                                       & \multicolumn{1}{c|}{58.7}                                       & \multicolumn{1}{c|}{51.7}                                       & \multicolumn{1}{c|}{45.9}                                       & 31.9                                       \\
CGDM \cite{du2021cross}                     & \multicolumn{1}{c||}{CVPR'21}      & \multicolumn{1}{c|}{43.7}                                       & \multicolumn{1}{c|}{40.5}                                       & \multicolumn{1}{c|}{36.1}                                        & \multicolumn{1}{c|}{33.0}                                        & \multicolumn{1}{c||}{23.4}                                       & \multicolumn{1}{c|}{60.5}                                       & \multicolumn{1}{c|}{57.3}                                       & \multicolumn{1}{c|}{54.1}                                       & \multicolumn{1}{c|}{\color[HTML]{3531ff} \textbf{50.6 (11.0↑)}}                          & \color[HTML]{3531ff} \textbf{40.9 (17.2↑)}                          \\ 
{SNSPGAN \cite{wang2022cycle}}                & \multicolumn{1}{c||}{ITS'22}                    
& \multicolumn{1}{c|}{51.0}                           & \multicolumn{1}{c|}{42.5}                      
& \multicolumn{1}{c|}{32.1}                           & \multicolumn{1}{c|}{28.4}                           
& \multicolumn{1}{c||}{15.3}                          & \multicolumn{1}{c|}{59.2}                      
& \multicolumn{1}{c|}{52.2}                           & \multicolumn{1}{c|}{41.7}                           
& \multicolumn{1}{c|}{36.6}                                       & {21.2}                                     
\\

D4 \cite{yang2022D4}                       & \multicolumn{1}{c||}{CVPR'22}                 & \multicolumn{1}{c|}{\color[HTML]{3531ff} \textbf{60.2 (6.6↑)}}                           & \multicolumn{1}{c|}{\color[HTML]{3531ff} \textbf{56.2 (10.1↑)}}                          & \multicolumn{1}{c|}{\color[HTML]{3531ff} \textbf{48.5 (11.8↑)}}                           & \multicolumn{1}{c|}{\color[HTML]{3531ff} \textbf{42.2 (10.6↑)}}                           & \multicolumn{1}{c||}{18.3}                                       & \multicolumn{1}{c|}{\color[HTML]{fe0000} \textbf{66.1 (4.7↑)}}                        & \multicolumn{1}{c|}{\color[HTML]{3531ff} \textbf{62.1 (6.8↑)}}                           & \multicolumn{1}{c|}{\color[HTML]{3531ff} \textbf{54.2 (8.0↑)}}                           & \multicolumn{1}{c|}{47.4}                                       & 24.3                                       \\

DAN-TransPar  \cite{han2022learning}            & \multicolumn{1}{c||}{TIP'22}          & \multicolumn{1}{c|}{23.2}                                       & \multicolumn{1}{c|}{17.7}                                       & \multicolumn{1}{c|}{13.2}                                        & \multicolumn{1}{c|}{11.4}                                        & \multicolumn{1}{c||}{5.90}                                       & \multicolumn{1}{c|}{40.6}                                       & \multicolumn{1}{c|}{42.7}                                       & \multicolumn{1}{c|}{29.2}                                       & \multicolumn{1}{c|}{35.0}                                       & 0.10                                       \\
MDD-TransPar \cite{han2022learning}             & \multicolumn{1}{c||}{TIP'22}    & \multicolumn{1}{c|}{20.8}                                       & \multicolumn{1}{c|}{16.7}                                       & \multicolumn{1}{c|}{13.1}                                        & \multicolumn{1}{c|}{13.0}                                        & \multicolumn{1}{c||}{8.60}                                       & \multicolumn{1}{c|}{53.4}                                       & \multicolumn{1}{c|}{39.9}                                       & \multicolumn{1}{c|}{35.5}                                       & \multicolumn{1}{c|}{33.8}                                       & 27.1                                       \\
SLP \cite{ling2023single}                     & \multicolumn{1}{c||}{TIP'23}    & \multicolumn{1}{c|}{58.9}                                       & \multicolumn{1}{c|}{55.2}                                       & \multicolumn{1}{c|}{47.8}                                        & \multicolumn{1}{c|}{39.8}                                        & \multicolumn{1}{c||}{24.2}                                       & \multicolumn{1}{c|}{64.5}                                       & \multicolumn{1}{c|}{60.7}                                       & \multicolumn{1}{c|}{53.2}                                       & \multicolumn{1}{c|}{45.6}                                       & 30.4                                       \\
C2PNet \cite{zheng2023curricular}                     & \multicolumn{1}{c||}{CVPR'23}    & \multicolumn{1}{c|}{54.9}                                       & \multicolumn{1}{c|}{48.2}                                       & \multicolumn{1}{c|}{37.6}                                        & \multicolumn{1}{c|}{30.7}                                        & \multicolumn{1}{c||}{16.1}                                       & \multicolumn{1}{c|}{62.5}                                       & \multicolumn{1}{c|}{56.4}                                       & \multicolumn{1}{c|}{46.1}                                       & \multicolumn{1}{c|}{38.7}                                       & 22.2                                       \\

\cellcolor[HTML]{EAEAEA}\textbf{UFEM (Ours)}               & \multicolumn{1}{c||}{\cellcolor[HTML]{EAEAEA}\textbf{-}}                    & \multicolumn{1}{c|}{\color[HTML]{fe0000} \cellcolor[HTML]{EAEAEA}\textbf{61.7 (8.1↑)}}                        & \multicolumn{1}{c|}{\color[HTML]{fe0000} \cellcolor[HTML]{EAEAEA}\textbf{59.2 (13.1↑)}}                       & \multicolumn{1}{c|}{\color[HTML]{fe0000} \cellcolor[HTML]{EAEAEA}\textbf{56.4 (19.7↑)}}                        & \multicolumn{1}{c|}{\color[HTML]{fe0000} \cellcolor[HTML]{EAEAEA}\textbf{53.2 (21.6↑)}}                        & \multicolumn{1}{c||}{\color[HTML]{fe0000} \cellcolor[HTML]{EAEAEA}\textbf{43.6 (26.1↑)}}                       & \multicolumn{1}{c|}{\color[HTML]{3531ff} \cellcolor[HTML]{EAEAEA}\textbf{64.4 (3.0↑)}}                           & \multicolumn{1}{c|}{\color[HTML]{fe0000} \cellcolor[HTML]{EAEAEA}\textbf{62.3 (7.0↑)}}                        & \multicolumn{1}{c|}{\color[HTML]{fe0000} \cellcolor[HTML]{EAEAEA}\textbf{57.1 (10.9↑)}}                       & \multicolumn{1}{c|}{\color[HTML]{fe0000} \cellcolor[HTML]{EAEAEA}\textbf{52.4 (12.8↑)}}                       & \color[HTML]{fe0000} \cellcolor[HTML]{EAEAEA}\textbf{42.0 (18.3↑)}                       \\  \hline


\multicolumn{12}{c}{\textbf{Degradation Type: Motion Blur} (\textbf{100} unpaired images for UFEM, \textbf{1200} unpaired images for UIR methods, \textbf{10K} unpaired images for UDA methods)}      \\ \hline
Baseline                 & \multicolumn{1}{c||}{-}                & \multicolumn{1}{c|}{55.0}                                       & \multicolumn{1}{c|}{40.7}                                       & \multicolumn{1}{c|}{24.0}                                        & \multicolumn{1}{c|}{12.7}                                        & \multicolumn{1}{c||}{8.30}                                       & \multicolumn{1}{c|}{64.5}                                       & \multicolumn{1}{c|}{53.9}                                       & \multicolumn{1}{c|}{37.5}                                       & \multicolumn{1}{c|}{21.9}                                       & 14.8                                       \\
{CycleGAN \cite{zhu2017unpaired}}                 & \multicolumn{1}{c||}{ICCV'17}                  & \multicolumn{1}{c|}{46.5}        & \multicolumn{1}{c|}{36.2}         & \multicolumn{1}{c|}{23.1}         & \multicolumn{1}{c|}{14.2}         & \multicolumn{1}{c||}{\color[HTML]{3531ff}{\textbf{11.6 (3.3↑)}}}  & \multicolumn{1}{c|}{57.8}        & \multicolumn{1}{c|}{49.2}        & \multicolumn{1}{c|}{36.0}         & \multicolumn{1}{c|}{24.5}         & {\color[HTML]{3531ff}{\textbf{20.1 (5.3↑)}}}\\

UID-GAN \cite{lu2019uid}                  & \multicolumn{1}{c||}{TBIOM'19}                     & \multicolumn{1}{c|}{39.0}                                       & \multicolumn{1}{c|}{22.3}                                       & \multicolumn{1}{c|}{8.10}                                        & \multicolumn{1}{c|}{3.60}                                        & \multicolumn{1}{c||}{2.30}                                       & \multicolumn{1}{c|}{50.1}                                       & \multicolumn{1}{c|}{33.5}                                       & \multicolumn{1}{c|}{13.6}                                       & \multicolumn{1}{c|}{5.90}                                       & 3.70                                       \\
DBGAN \cite{zhang2020deblurring}                   & \multicolumn{1}{c||}{CVPR'20}                 & \multicolumn{1}{c|}{{\color[HTML]{3531ff} \textbf{55.6 (0.6↑)}}}    & \multicolumn{1}{c|}{{\color[HTML]{3531ff} \textbf{44.7 (4.0↑)}}}    & \multicolumn{1}{c|}{{\color[HTML]{3531ff} \textbf{28.6 (4.6↑)}}}     & \multicolumn{1}{c|}{{\color[HTML]{3531ff} \textbf{15.4 (2.7↑)}}}     & \multicolumn{1}{c||}{\color[HTML]{000000} 10.0}                           & \multicolumn{1}{c|}{63.9}                                       & \multicolumn{1}{c|}{{\color[HTML]{3531ff} \textbf{56.8 (2.9↑)}}}    & \multicolumn{1}{c|}{{\color[HTML]{3531ff} \textbf{42.4 (4.9↑)}}}    & \multicolumn{1}{c|}{{\color[HTML]{3531ff} \textbf{25.3 (3.4↑)}}} & 17.1                                       \\
CGDM \cite{du2021cross}                     & \multicolumn{1}{c||}{CVPR'21}                                     & \multicolumn{1}{c|}{{\color[HTML]{000000} 39.7}}                & \multicolumn{1}{c|}{{\color[HTML]{000000} 28.6}}                & \multicolumn{1}{c|}{{\color[HTML]{000000} 14.8}}                 & \multicolumn{1}{c|}{3.00}                                        & \multicolumn{1}{c||}{1.80}                                       & \multicolumn{1}{c|}{{\color[HTML]{000000} 60.3}}                & \multicolumn{1}{c|}{{\color[HTML]{000000} 52.9}}                & \multicolumn{1}{c|}{{\color[HTML]{000000} 40.8}}                & \multicolumn{1}{c|}{{\color[HTML]{000000} 24.7}}    & {\color[HTML]{000000} 19.2}    \\ 

DAN-TransPar \cite{han2022learning}             & \multicolumn{1}{c||}{TIP'22}           & \multicolumn{1}{c|}{29.1}                                       & \multicolumn{1}{c|}{17.2}                                       & \multicolumn{1}{c|}{8.60}                                        & \multicolumn{1}{c|}{4.50}                                        & \multicolumn{1}{c||}{3.30}                                       & \multicolumn{1}{c|}{44.6}                                       & \multicolumn{1}{c|}{30.1}                                       & \multicolumn{1}{c|}{0.10}                                       & \multicolumn{1}{c|}{0.70}                                       & 9.30                                       \\
MDD-TransPar \cite{han2022learning}            & \multicolumn{1}{c||}{TIP'22}         & \multicolumn{1}{c|}{29.0}                                       & \multicolumn{1}{c|}{13.6}                                       & \multicolumn{1}{c|}{8.10}                                        & \multicolumn{1}{c|}{4.20}                                        & \multicolumn{1}{c||}{3.10}                                       & \multicolumn{1}{c|}{56.7}                                       & \multicolumn{1}{c|}{42.9}                                       & \multicolumn{1}{c|}{26.8}                                       & \multicolumn{1}{c|}{16.3}                                       & 12.0                                       \\
FCL-GAN \cite{zhao2022fcl}                 & \multicolumn{1}{c||}{ACM MM'22}            & \multicolumn{1}{c|}{54.5}                                       & \multicolumn{1}{c|}{42.2}                                       & \multicolumn{1}{c|}{25.7}                                        & \multicolumn{1}{c|}{13.5}                                        & \multicolumn{1}{c||}{5.70}                                       & \multicolumn{1}{c|}{{\color[HTML]{3531ff} \textbf{64.9 (0.4↑)}}}    & \multicolumn{1}{c|}{55.7}                                       & \multicolumn{1}{c|}{40.7}                                       & \multicolumn{1}{c|}{24.9}                                       & 13.1                                       \\
CRNet \cite{zhao2022crnet}                  & \multicolumn{1}{c||}{ACM MM'22}                   & \multicolumn{1}{c|}{22.9}                                       & \multicolumn{1}{c|}{17.6}                                       & \multicolumn{1}{c|}{6.40}                                        & \multicolumn{1}{c|}{2.90}                                        & \multicolumn{1}{c||}{3.90}                                       & \multicolumn{1}{c|}{32.7}                                       & \multicolumn{1}{c|}{27.2}                                       & \multicolumn{1}{c|}{11.9}                                       & \multicolumn{1}{c|}{5.70}                                       & 6.60                                       \\               

\cellcolor[HTML]{EAEAEA}\textbf{UFEM (Ours)}                & \multicolumn{1}{c||}{\cellcolor[HTML]{EAEAEA}\textbf{-}}           & \multicolumn{1}{c|}{{\color[HTML]{fe0000} \cellcolor[HTML]{EAEAEA}\textbf{59.7 (4.7↑)}}} & \multicolumn{1}{c|}{{\color[HTML]{fe0000} \cellcolor[HTML]{EAEAEA}\textbf{50.6 (9.9↑)}}} & \multicolumn{1}{c|}{{\color[HTML]{fe0000} \cellcolor[HTML]{EAEAEA}\textbf{34.7 (10.7↑)}}} & \multicolumn{1}{c|}{{\color[HTML]{fe0000} \cellcolor[HTML]{EAEAEA}\textbf{22.8 (10.1↑)}}} & \multicolumn{1}{c||}{{\color[HTML]{fe0000} \cellcolor[HTML]{EAEAEA}\textbf{15.3 (7.0↑)}}} & \multicolumn{1}{c|}{{\color[HTML]{fe0000} \cellcolor[HTML]{EAEAEA}\textbf{65.7 (1.2↑)}}} & \multicolumn{1}{c|}{{\color[HTML]{fe0000} \cellcolor[HTML]{EAEAEA}\textbf{58.2 (4.3↑)}}} & \multicolumn{1}{c|}{{\color[HTML]{fe0000} \cellcolor[HTML]{EAEAEA}\textbf{43.1 (5.6↑)}}} & \multicolumn{1}{c|}{{\color[HTML]{fe0000} \cellcolor[HTML]{EAEAEA}\textbf{26.7 (4.8↑)}}}      & {\color[HTML]{fe0000} \cellcolor[HTML]{EAEAEA}\cellcolor[HTML]{EAEAEA}\textbf{20.1 (5.3↑)}}  \\
\Xhline{1.5pt}

\end{tabular}}
\end{table*}

\subsection{Low-Quality Image Classification}
\textbf{Quantitative Results.}
To fully verify the effectiveness and generality of our UFEM, we compare it with 20 kinds of IR and UDA methods. We conduct the evaluation on one large-scale synthetic dataset (\emph{i.e.}, ImageNet-C \cite{hendrycks2019benchmarking}) and two real degradation datasets(\emph{i.e.}, Haze-20 \cite{pei2018does} and ExDARK \cite{loh2019getting}). For ImageNet-C, we choose fog and motion blur for evaluation. As can be seen from Tab. \ref{tab:cls_syn} and Tab. \ref{tab:cls_real}, our proposed method significantly surpasses the IR and UDA methods in terms of classification accuracy on synthetic and real degradation datasets. The UDA methods as the general generalization means, struggle to discern specific degradation types, leading to universally lower performance compared to IR methods. In addition, since they rely on unpaired but semantic-aligned images for retraining, which are not available in real-world applications, we mainly focus on comparison with the IR methods on real datasets. Continuing, when it comes to real foggy and dark conditions, the improvement of IR methods is also limited, while our method can improve the accuracy of pre-trained networks significantly. For example, on the Haze-20 dataset, the best-performing D4 \cite{yang2022D4} improves ResNet50 by only 0.3\%, while our method can boost by 9.4\%. Similar conclusions can be found on the ExDARK dataset. All this demonstrates that our UFEM effectively boosts the performance of pre-trained models, beneficial for autonomous vehicles to improve visual perception and navigational safety under real degradation conditions.
\begin{table}[htbp]
\centering
\scriptsize
\caption{Quantitative comparison of our UFEM with IR methods on real foggy and real dark datasets for image recognition.}
\label{tab:cls_real}
\scalebox{0.94}{
\begin{tabular}{l|cc||cc}
\Xhline{1.5pt}
                                      &                                            &                                  & \multicolumn{2}{c}{\textbf{\emph{Top-1 Acc (\%)}}}      
                                      \\
\multirow{-2}{*}{$\mathbb{D}$  }                    & \multirow{-2}{*}{\textbf{Method}}          & \multirow{-2}{*}{\textbf{Venue}} & \textbf{VGG16}                        & \textbf{ResNet50}                     \\ \hline
                                       & Baseline                                   & -                         & 49.6                                  & 42.8                                  \\
                                       & YOLY \cite{li2021you}                                       & IJCV'21                   & 47.0                                  & 42.1                                  \\
                                       & RefineDNet \cite{zhao2021refinednet}                                & TIP'21                    & 47.5                                  & 40.0                                  \\
                                       & SNSPGAN \cite{wang2022cycle}                                   & ITS'22                    & 50.4                                  & 42.5                                     \\
                                       & D4 \cite{yang2022D4}                                         & CVPR'22                   & \color[HTML]{3531FF} \textbf{50.5 (0.9↑)}                                  & \color[HTML]{3531FF} \textbf{43.1 (0.3↑)}                                  \\     
                                       & SLP \cite{ling2023single}                                       & TIP'23                    & 50.4                                  & 43.0                                  \\ 
                                       & C2PNet \cite{zheng2023curricular}                               & CVPR'23                    & 50.0                                  & 42.6                                     \\ 
\multirow{-7}{*}{\rotatebox{90}{\textbf{Haze-20 \cite{pei2018does}}}} & \cellcolor[HTML]{EAEAEA}\textbf{UFEM (Ours)} & \cellcolor[HTML]{EAEAEA}- & \color[HTML]{fe0000} \cellcolor[HTML]{EAEAEA}\textbf{51.2 (1.6↑)} &\color[HTML]{fe0000} \cellcolor[HTML]{EAEAEA}\textbf{52.2 (9.4↑)} \\ \hline
                                       & Baseline                                   & -                         & 49.5                                  & 46.3                                  \\
                                       & Zero-DCE \cite{guo2020zero}                                  & CVPR'20                   & 46.6                                  & 44.8                                  \\
                                       & EnGAN \cite{jiang2021enlightengan}                                     & TIP'21                    &\color[HTML]{3531ff} \textbf{50.6 (1.1↑)}                                  & 48.6                                 \\
                                       & Zero-DCE++ \cite{li2021learning}                                & TPAMI'21                  & 47.2                                  & 46.7                                  \\
                                       & LE-GAN \cite{fu2022gan}                                    & KBS'22                    & 38.4                                  & 35.4                                  \\
                                       & NeRCo \cite{yang2023implicit}                                      & ICCV'23                   & 36.0                                     & 38.6                                     \\ 
                                       & CUE \cite{zheng2023empowering}                                      & ICCV'23                   & 48.7                                     & \color[HTML]{3531ff} \textbf{52.8 (6.5↑)}                                     \\ 
\multirow{-7}{*}{\rotatebox{90}{\textbf{ExDARK \cite{loh2019getting}}}}  & \cellcolor[HTML]{EAEAEA}\textbf{UFEM (Ours)} & \cellcolor[HTML]{EAEAEA}- & \color[HTML]{fe0000} \cellcolor[HTML]{EAEAEA}\textbf{52.4 (2.9↑)} &\color[HTML]{fe0000} \cellcolor[HTML]{EAEAEA}\textbf{54.2 (7.9)↑} \\ \Xhline{1.5pt}
\end{tabular}}
\end{table}

\begin{figure}[htbp]
\centering
\includegraphics[width=0.92\linewidth]{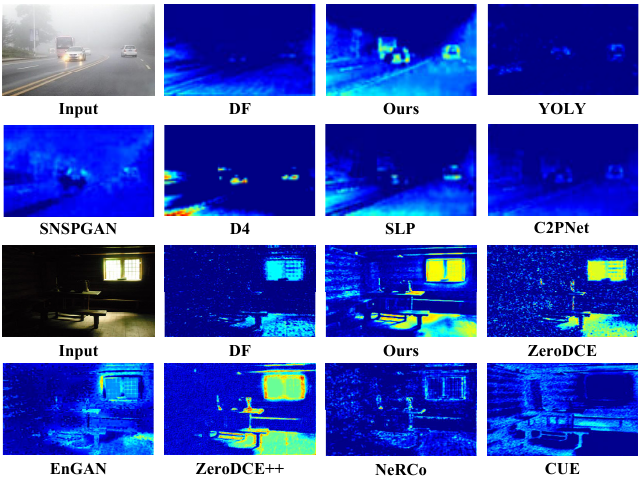}
\caption{The visualization comparison of enhanced features generated by the top 5 UIR methods and our UFEM, respectively.}
\label{fig:features}
\end{figure}

\textbf{Feature Visualization.} To demonstrate the performance of our UFEM in feature correction, we visualize the feature maps of real foggy and dark images from different methods, as shown in Fig. \ref{fig:features}. It can be seen that degradation cues will lead to varying degrees of corruption in the discriminative structures of features, causing content loss and introducing additional artifacts, \emph{e.g.}, cars vanishing in fog, and rooms submerged in dark condition. While IR methods can restore certain structural information in images, they do not consistently enhance the discriminative feature responses of objects. For instance, YOLY 
\cite{li2021you} and D4 \cite{yang2022D4} can only partially enhance local areas in features, leaving other areas still in haze. Besides, ZeroDCE \cite{guo2020zero} and NeRCo \cite{yang2023implicit} introduce more noise to the dark-shrouded areas, \emph{i.e.}, artifacts. Although the superior CUE \cite{zheng2023empowering} can outline the room, its details are still lost, \emph{e.g.}, the kettles and cups on the table. In contrast, our method significantly boosts the feature responses in the discriminative regions of images, not only achieving effective restoration of latent content but also suppressing additional artifacts.

\begin{figure}[htbp]
\centering
\includegraphics[width=0.96\linewidth]{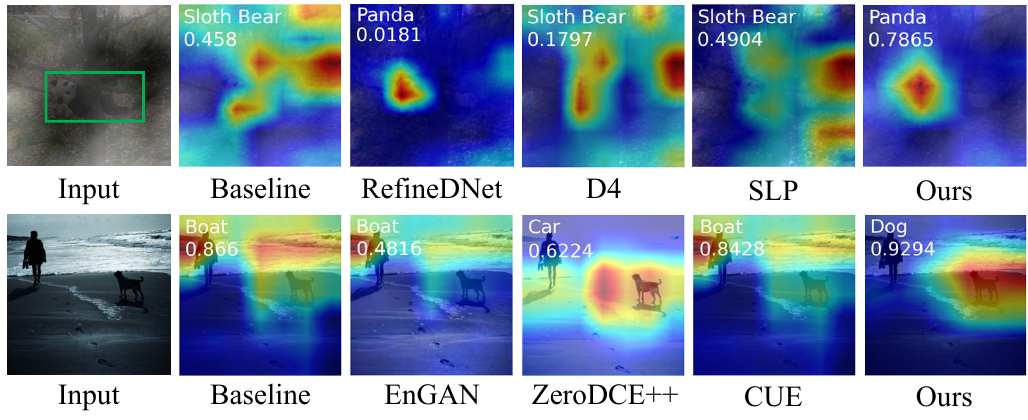}
\caption{Grad-CAM attention maps comparison of our UFEM with UIR methods. We choose the Top-3 IR methods for visualization. Top-1 prediction of the classifier ResNet50 together with its confidence score is presented.}
\label{fig:gradcam2}
\end{figure}

\textbf{Grad-CAM Attention Comparison.} Besides, we compare the contribution of our UFEM and UIR methods in accurately drawing the attention of classifiers leveraging grad-CAMs \cite{selvaraju2017grad}. As shown in Fig. \ref{fig:gradcam2}, most UIR methods fail to effectively guide the classifier to focus more accurately on the discriminative regions of the image. This, in turn, results in misclassifications by the classifier, such as the inability to correctly identify a panda in a tree, and a dog by the riverside after image-level enhancement. While certain UIR methods (\emph{e.g.}, RefineDNet \cite{zhao2021refinednet}) enable the classifiers to correctly identify objects in the enhanced images, their prediction scores are often low, indicating a lack of confidence when dealing with enhanced images as inputs. In contrast, our UFEM better enhances the discriminative regions around objects, enabling focused attention and accurate prediction, thereby benefiting deep network recognition.

\begin{figure}[htbp]
\centering
\includegraphics[width=0.96\linewidth]{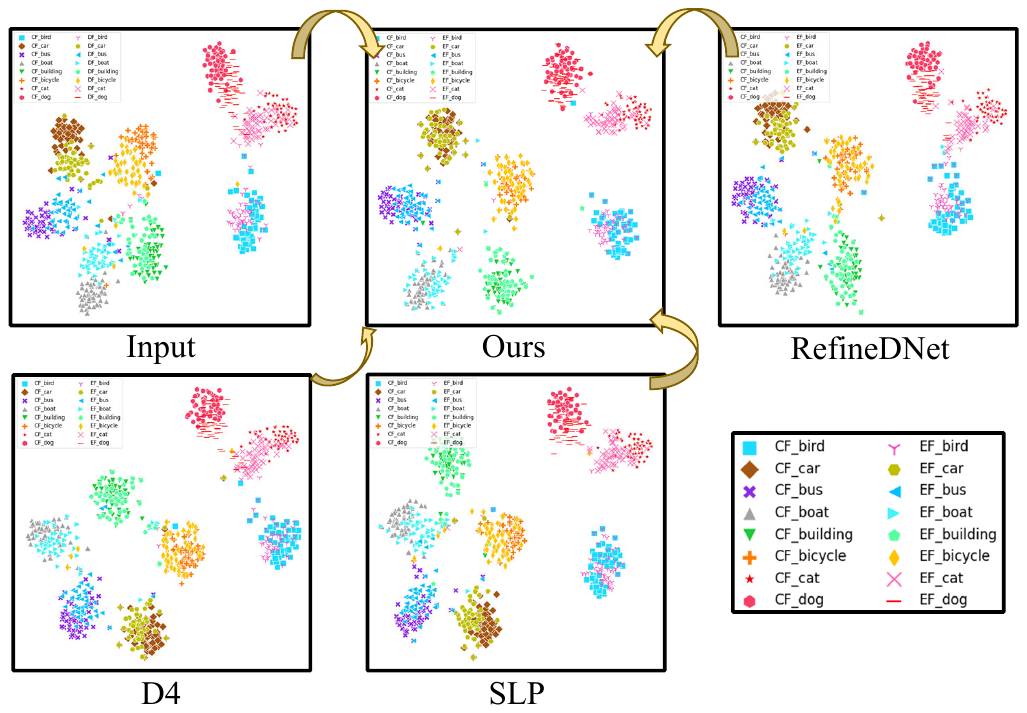}
\caption{T-SNE visual comparison of feature embeddings from VGG16 with our UFEM and Top-3 UIR methods. Zoom in for the best view.}
\label{fig:tsne}
\end{figure}

\textbf{T-SNE Visualization.} Finally, we leverage t-SNE \cite{van2008visualizing} to visualize the distributions of degraded, clear, and enhanced features to verify the effectiveness of our method in bridging the gap between degraded features and clear ones. Firstly, we present the feature distribution comparisons with other IR methods. Specifically, we select the 8 most common categories from ImageNet and Fog-4 as inputs, and extract the final layer output of VGG16 as feature embeddings for visualization. As shown in Fig. \ref{fig:tsne}, due to the interference of degradation cues, there is an obvious domain drift between degraded and clear features. Indeed, after enhancement with D4 \cite{yang2022D4} and SLP \cite{ling2023single}, some degree of bridging the gap between degraded and clean domains is achieved. However, the domain shift still persists. In contrast, following enhancement by our UFEM, the distribution between enhanced and clear features is effectively aligned, improving the performance of existing models on degraded images. 

\begin{figure}[htbp]
\centering
\includegraphics[width=0.96\linewidth]{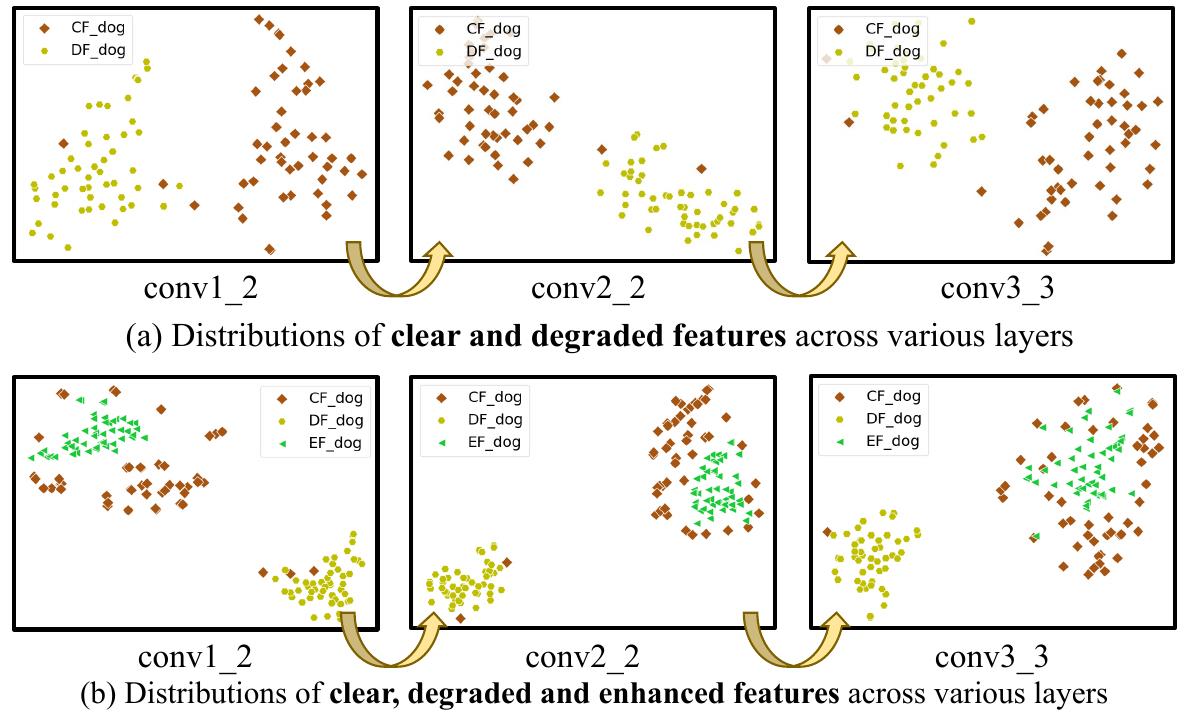}
\caption{T-SNE visualization of features across different network layers.}
\label{fig:tsne2}
\end{figure}

Apart from the above, we also provide the feature distribution comparisons across different layers. Specifically, we focus on the “dog” form ImageNet-C, and leverage the “conv1\_2”, “conv2\_2”, and “conv3\_3” of VGG16 to extract three types of features for t-SNE visualization. As depicted in Fig. \ref{fig:tsne2}(a), due to the influence of degradation cues, there is an obvious gap between clear and degraded features in shallow layers, and as network layers deepen, the disparities continuously propagate, causing a similar domain shift in deep layers, and ultimately leading to erroneous recognition. In contrast, as shown in Fig. \ref{fig:tsne2}(b), after our UFEM enhancement, we effectively aligned the distributions between enhanced and clear features, thereby reducing error accumulation in forward propagation. After that, the consistent distributions of enhanced features with clear ones across various layers were attained, and ultimately benefiting deep network recognition.

\begin{table}[htbp]
\scriptsize
\centering
\caption{Quantitative comparison of our UFEM with IR methods on two real foggy datasets for object detection.}
\label{tab:det}
\renewcommand\arraystretch{1.1}
\scalebox{0.9}{
\begin{tabular}{l||cc||cc}
\Xhline{1.5pt}
                          & \multicolumn{2}{c||}{\textbf{\emph{RTTS (\%)}}}                              & \multicolumn{2}{c}{\textbf{\emph{DAWN (\%)}}}                         \\ 
\multirow{-2}{*}{\textbf{Methods}} & \textbf{mAP50}          & \textbf{mAP}            & \textbf{mAP50}          & \textbf{mAP}            \\ 
\hline
Baseline                  & 23.7       & 12.9          & {9.9}    & 5.9          \\
YOLY \cite{li2021you}                     & 19.5       & 8.37          & 11.5          & 4.9          \\
RefineDNet \cite{zhao2021refinednet}               & 26.7       & 14.5          & 14.3          & 7.7    \\
SNSPGAN \cite{wang2022cycle}                  & 24.8       & 13.5          & 14.9          & 8.7              \\
D4 \cite{yang2022D4}                       & 27.8       & 15.0    & 15.6          & 8.5          \\
SLP \cite{ling2023single}                      & \color[HTML]{3531ff} \textbf{27.8 (4.1↑)}       & \color[HTML]{3531ff} \textbf{15.1 (2.2↑)}     & \color[HTML]{3531ff} \textbf{16.6 (6.7↑)}          & \color[HTML]{3531ff} \textbf{9.1 (3.2↑)}          \\
C2PNet \cite{zheng2023curricular}                       & 19.2      & 8.19    & 11.8          & 6.9          \\
\cellcolor[HTML]{EAEAEA}\textbf{UFEM (Ours)}                & \cellcolor[HTML]{EAEAEA}\color[HTML]{fe0000}\textbf{28.8 (5.1↑)} & \cellcolor[HTML]{EAEAEA}\color[HTML]{fe0000}\textbf{15.7 (2.8↑)}    & \cellcolor[HTML]{EAEAEA}\color[HTML]{fe0000}\textbf{17.5 (7.6↑)} & \cellcolor[HTML]{EAEAEA}\color[HTML]{fe0000}\textbf{9.3 (3.4↑)} \\

\Xhline{1.5pt}

\end{tabular}}
\end{table}

\subsection{Foggy Object Detection}
\textbf{Quantitative Results.}
Compared to image recognition, object detection requires both identification and spatial localization of objects, demanding greater completeness and richness in feature content. To further validate the effectiveness of our UFEM in feature correction, we conduct experimental verification on two real foggy object detection datasets (\emph{i.e.}, RTTS \cite{li2018benchmarking} and DAWN \cite{hassaballah2020vehicle}) and compare it with six SOTA IR methods. As reported in Tab. \ref{tab:det}, our method exhibits consistent improvements in performance across two real foggy scenarios, outperforming all IR methods. This demonstrates that our UFEM can more accurately restore the content and details of degraded features, which, in turn, facilitates improved object recognition and localization within detection networks and stronger safety of autonomous driving.

\begin{figure}[htbp]
\centering
\includegraphics[width=0.94\linewidth]{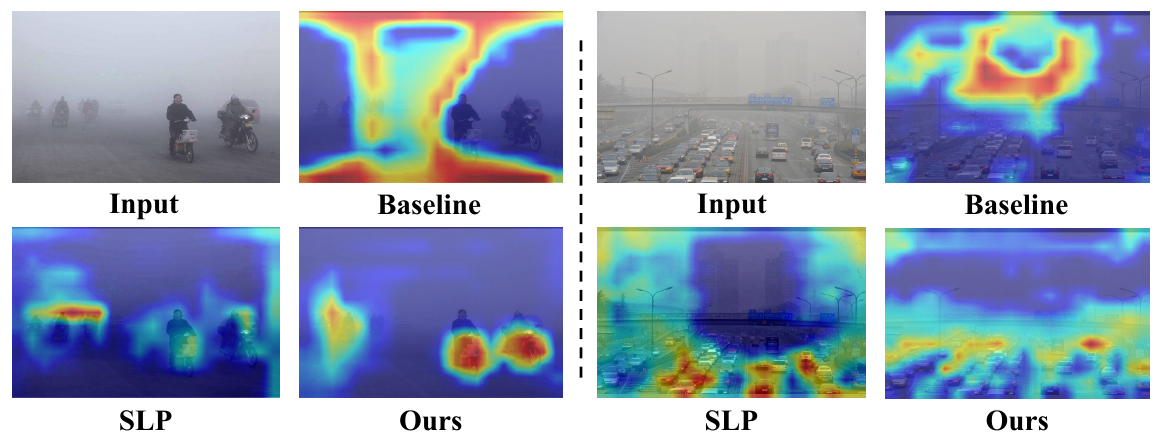}
\caption{Grad-CAM attention maps comparison of UFEM with UIR methods. }
\label{fig:gradcam}
\end{figure}

\textbf{Grad-CAM Attention Visualization.} We next compare the contribution of our UFEM and the best-performing IR method in accurately drawing the attention of detector leveraging gradCAMs \cite{selvaraju2017grad}. As shown in Fig. \ref{fig:gradcam}, SLP \cite{ling2023single} sometimes fail to effectively guide the detector to focus more accurately on the discriminative regions of the image. This, in turn, leads to missed and false detections by the detector, such as the inability to accurately detect the looming riders in fog and the erroneous focus on background areas like the sky. In contrast, our UFEM better enhances the discriminative regions surrounding the objects, allowing the detector to attentively focus and predict the correct category and spatial location of objects. This demonstrates the recognition-friendly property of our UFEM for deep neural networks.

\begin{figure}[htbp]
\centering
\includegraphics[width=0.92\linewidth]{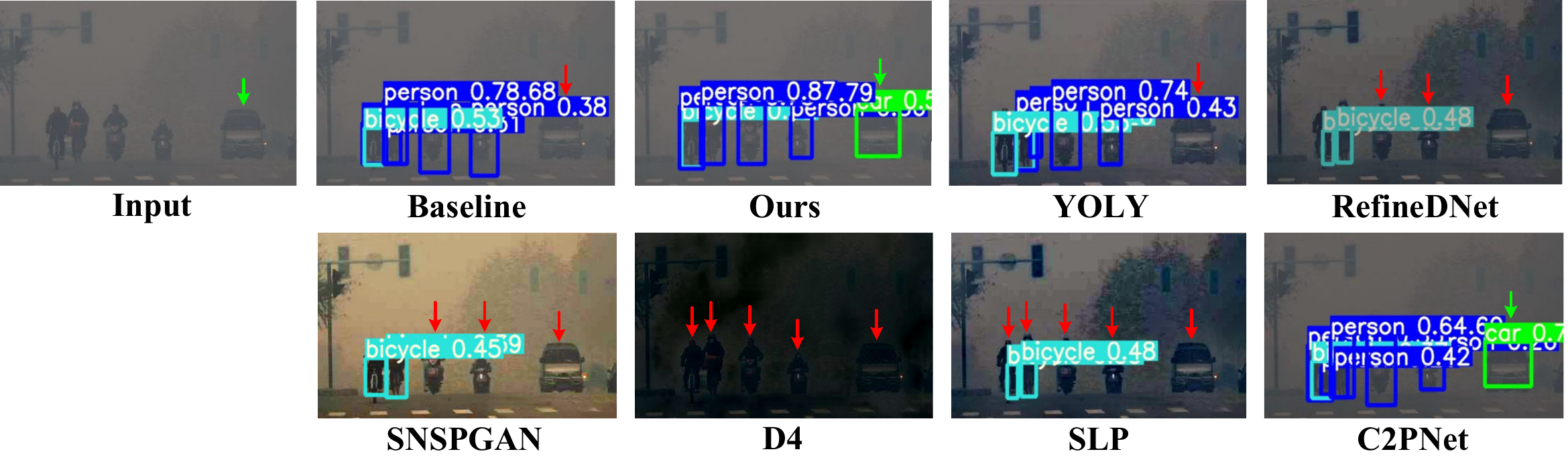}
\caption{Object detection results comparison of our UFEM with UIR methods.}
\label{fig:detection}
\end{figure}

\textbf{Object Detection Comparison.} We further present visual comparisons of object detection results with YOLOv5 \cite{redmon2016you}. As shown in Fig. \ref{fig:detection}, after processing with IR methods, feeding the restored images into the detector occasionally yields positive detection results, as indicated by the green arrows in the images. However, in the majority of cases, it leads to worse detections, as indicated by the red arrows in the images. In certain instances, particularly following D4 \cite{yang2022D4} enhancement, the detector exhibits a tendency to overlook all objects within the image. In contrast, our UFEM notably mitigates false negatives and false positives in environmental perception, such as cars driving in the haze, thus ensuring the safety of vehicle navigation in real degradation scenarios.


\subsection{Nighttime Semantic Segmentation}
\textbf{Quantitative Results.}
Semantic segmentation relies on the local consistency of the image to perform pixel-level classification for each point \cite{chen2018encoder}. However, the introduction of degradation cues will disrupt the image's statistical and structural characteristics, posing challenges to pre-trained networks for achieving high-precision semantic segmentation. To confirm the accuracy of UFEM in local feature content restoration, we conduct experiments on three real dark segmentation datasets (\emph{i.e.}, ACDC Nighttime \cite{sakaridis2021acdc}, Dark\_Zurich \cite{sakaridis2019guided}, and Nighttime Driving \cite{dai2018dark}), and compare it with six advanced IR methods. As reported in Tab. \ref{tab:seg}, a remarkable observation is that all IR-based methods suffer a significant degradation in performance, rather than improvement. One potential reason is that existing IR models primarily focus on lower resolutions for restoration \cite{jiang2021enlightengan}\cite{guo2020zero}, like 512×512. They may perform poorly on high-resolution nighttime images, like 1920×1080, resulting in reduced accuracy. In contrast, our method achieves excellent feature correction across various resolutions, enhancing performance on all three real datasets and exceeding IR methods by \textbf{7.5$\sim$10.1\%}. This further validates the effectiveness and generality of our UFEM.

\begin{figure}[htbp]
\centering
\includegraphics[width=0.92\linewidth]{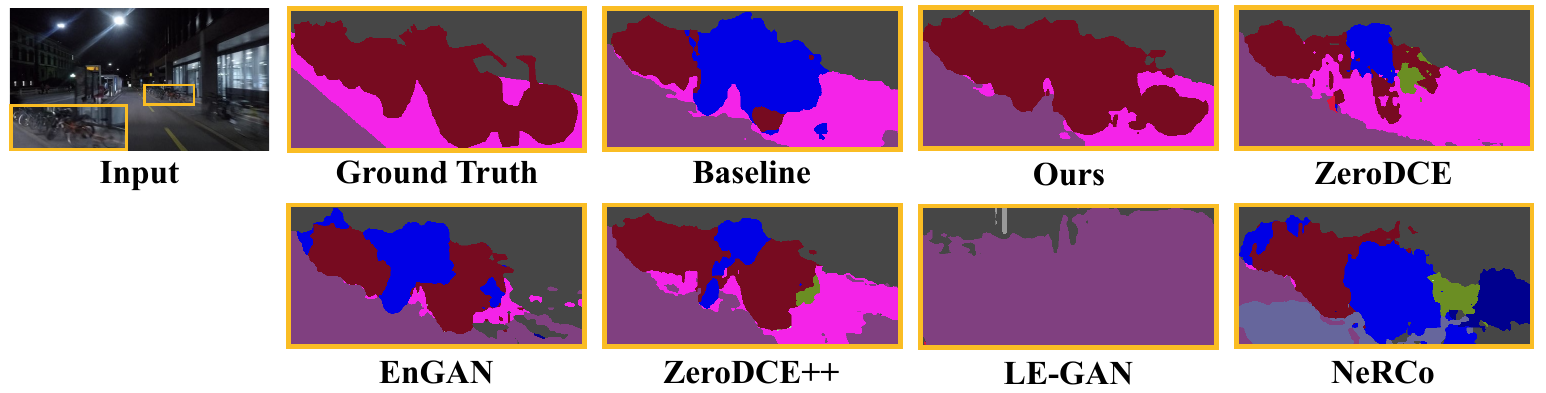}
\caption{Segmentation results comparison of our UFEM with UIR methods.}
\label{fig:segmentation}
\end{figure}

\textbf{Semantic Segmentation Visualization.} We finally present a qualitative comparison between our UFEM and IR methods. As shown in Fig. \ref{fig:segmentation}, our method obviously generates more accurate segmentation results for both distant and dense objects in the dark, such as a row of bicycles parked side by side. In conclusion, all the results affirm the effectiveness of our UFEM across various degradation scenarios and tasks, demonstrating its capability to endow autonomous vehicles with high-efficiency and high-robustness visual perception in real degradation conditions.

\begin{table}[htbp]
    \centering
    \footnotesize
    \caption{Quantitative comparison of our UFEM with IR methods on three nighttime segmentation datasets. {\color[HTML]{f97d1c}\textbf{Bold}} indicates the least performance decline, while {\color[HTML]{FE0000}\textbf{bold}} is same as above.}
    \label{tab:seg}
    \renewcommand\arraystretch{1.1}
    \scalebox{0.75}{
    \begin{tabular}{l||ccc}
    \Xhline{1.5pt}
      & \textbf{\emph{ACDC (night)}}                      & \textbf{\emph{Dark\_Zurich}}                      & \textbf{\emph{Nighttime Driving}}                 \\
\multirow{-2}{*}{\textbf{Method}}         & mIoU(\%)                                   & mIoU(\%)                                   & mIoU(\%)                                   \\ \hline
                                                
    Baseline                         & 22.1                                       & 15.7                                       & 28.9                                       \\
    ZeroDCE \cite{guo2020zero}                          & 14.0                                       & 10.4                                       & 19.0                                       \\
    EnGAN \cite{jiang2021enlightengan}                            & 15.5                                       & 11.4                                       & {\color[HTML]{f97d1c} \textbf{22.9 (6.0↓)}} \\
    ZeroDCE++ \cite{li2021learning}                        & 16.2                                       & {\color[HTML]{f97d1c} \textbf{13.0 (2.7↓)}} & 21.0                                       \\
    LE-GAN \cite{fu2022gan}                           & 16.0                                       & 12.6                                       & 21.4                                       \\
    NeRCo \cite{yang2023implicit}                            & {\color[HTML]{f97d1c} \textbf{16.6 (5.5↓)}} & 11.8                                       & 20.2                                       \\
    CUE \cite{zheng2023empowering}                       & {16.2} & 11.3                                       & 19.6                                       \\
    \rowcolor[HTML]{C0C0C0} 
    {\cellcolor[HTML]{EAEAEA} \textbf{UFEM (Ours)}} & {\cellcolor[HTML]{EAEAEA} \color[HTML]{FE0000}\textbf{26.7 (4.6↑)}} & {\cellcolor[HTML]{EAEAEA} \color[HTML]{FE0000}\textbf{20.5 (4.8↑)}} & {\cellcolor[HTML]{EAEAEA} \color[HTML]{FE0000}\textbf{32.3 (3.4↑)}} \\ \Xhline{1.5pt}
    \end{tabular}}
\end{table}

\subsection{Ablation Study}
\label{ablation}
\textbf{Effectiveness of Multi-adversarial Mechanism.} Firstly, to verify the benefits of multi-adversarial mechanism in Stage-1, we design several variants in Tab. \ref{tab:multidis} and \ref{tab:multidis2}. Among them, “1D”, “2D”, and “3D” refer to different numbers of discriminators in DPL, “V”, “R”, and “Y” respectively refer to VGG16, ResNet50, and YOLOv5. As we can see, equipping deep networks with more discriminators tends to yield better performance than the baseline, both on synthetic and real datasets, such as a 2.9\% improvement for UFEM\_3D to UFEM\_1D on ExDARK. Although the gains diminish with an increasing number of discriminators, approaching saturation, the multi-scale discriminators continue to be positive, such as UFEM on RTTS. Overall, these results affirm the positive role of the multi-adversarial mechanism in Stage-1 for enhancing the robustness and generalization of our UFEM. However, more discriminators also bring heavy parameters and computational complexity. For the trade-off between performance and computational overhead, we adopt the “3D” for multi-scale discrimination, effectively addressing the challenges posed by high spatial sparsity and preventing the error accumulation in feature forward propagation.

\begin{table}[htbp]
\footnotesize
\centering
\caption{The different numbers of discriminators used in Stage-1.}
\label{tab:multidis}
\renewcommand\arraystretch{1.2}
\scalebox{0.9}{
\begin{tabular}{l||ccc}
\Xhline{1.2pt}
                    & \multicolumn{3}{c}{\textbf{\emph{ImageNet-C (\%)}}} \\
\multirow{-2}{*}{\textbf{Method}}     & \textbf{fog1}         & \textbf{fog3}     & \textbf{fog5}   \\ \hline
VGG16                    & 53.6                 & 36.7          & 17.5       \\
V UFEM\_1D          & 59.1 (5.5↑)           & 48.8 (12.1↑)       & 32.1 (14.6↑) \\
V UFEM\_2D          & 59.2 (0.1↑)           & 49.7 (0.9↑)      & 34.0 (1.9↑) \\
\cellcolor[HTML]{EAEAEA}\textbf{V UFEM\_3D}          & \cellcolor[HTML]{EAEAEA}\textbf{59.8 (0.6↑)}           & \cellcolor[HTML]{EAEAEA}\textbf{51.1 (1.4↑)}      & \cellcolor[HTML]{EAEAEA}\textbf{34.5 (0.5↑)} \\
\Xhline{1.2pt}
\end{tabular}}
\end{table}

\begin{table}[htbp]
\footnotesize
\centering
\caption{The different numbers of discriminators used in Stage-1.}
\label{tab:multidis2}
\renewcommand\arraystretch{1.2}
\scalebox{0.9}{
\begin{tabular}{l||cccc}
\Xhline{1.2pt}
                    & \multicolumn{4}{c}{\textbf{\emph{UFEM\_Multi-adversarial (\%)}}} \\
\multirow{-2}{*}{\textbf{Datasets}}     & \textbf{Baseline}       & \textbf{1D}         & \textbf{2D}     & \cellcolor[HTML]{EAEAEA}\textbf{3D}   \\ \hline
\textbf{ExDARK (R)}        & 46.3    & 49.1 (2.8↑)           & 50.9 (1.8↑)     & \cellcolor[HTML]{EAEAEA}\textbf{52.0 (1.1↑)}    \\
\textbf{RTTS (Y)}       & 23.7    & 27.9 (4.2↑)          & 28.2 (0.3↑)         & \cellcolor[HTML]{EAEAEA}\textbf{28.4 (0.1↑)}    \\
\Xhline{1.2pt}
\end{tabular}}
\end{table}


\textbf{Effectiveness of Two-stage Correction.}  We next conduct an ablation to analyze the contribution of our two-stage correction strategy. As reported in Tab. \ref{tab:twostage} and \ref{tab:twostage2}, using only Stage-1 (S1) or Stage-2 (S2) shows some improvements as expected, with the performance of S1 surpassing S2 in most cases. This is because, in the case where the input features are severely degraded, prioritizing feature content restoration aids network recognition, indicating the necessity of our S1. However, when the input features are not severely degraded, such as with ResNet on ExDARK, S2 achieves slightly higher accuracy compared to S1.  Apart from the above, both S1 and S2 always fall significantly short of our two-stage method, \emph{i.e.}, S1+S2, indicating that sequentially performing content restoration and correlation modulation guided by DCP is more beneficial for feature correction. That is, our two-stage feature correction is effective, both on synthetic and real datasets.


\begin{table}[htbp]
\small
\centering
\caption{The necessity of two-stage correction is verified on Fog with two models, \emph{i.e.}, “V” for VGG16 and “R” for ResNet50.}
\renewcommand\arraystretch{1.1}
\label{tab:twostage}
\scalebox{0.84}{
\begin{tabular}{l||ccc}
\Xhline{1.2pt}
                    & \multicolumn{3}{c}{\textbf{\emph{ImageNet-C (\%)}}} \\
\multirow{-2}{*}{\textbf{Method}}     & \textbf{fog1}         & \textbf{fog3}     & \textbf{fog5}   \\ \hline
VGG16                    & 53.6                 & 36.7          & 17.5       \\
V UFEM (S1)          & 59.8 (6.2↑)           & 51.1 (14.4↑)       & 34.5 (17.0↑) \\
V UFEM (S2)          & 56.8 (3.2↑)           & 42.7 (6.0↑)      & 22.5 (5.0↑) \\
\cellcolor[HTML]{EAEAEA}\textbf{V UFEM (S1+S2)}          & \cellcolor[HTML]{EAEAEA}\textbf{61.7 (8.1↑)}           & \cellcolor[HTML]{EAEAEA}\textbf{56.4 (19.7↑)}      & \cellcolor[HTML]{EAEAEA}\textbf{43.6 (26.1↑)} \\ \hline

ResNet50                    & 61.4                 & 46.2          & 23.7       \\
R UFEM (S1)          & 63.8 (2.4↑)           & 54.8 (8.6↑)       & 38.4 (14.7↑) \\
R UFEM (S2)          & 63.5 (2.1↑)           & 52.9 (6.7↑)      & 35.4 (11.7↑) \\
\cellcolor[HTML]{EAEAEA}\textbf{R UFEM (S1+S2)}          & \cellcolor[HTML]{EAEAEA}\textbf{64.4 (3.0↑)}           & \cellcolor[HTML]{EAEAEA}\textbf{57.1 (10.9↑)}      & \cellcolor[HTML]{EAEAEA}\textbf{42.0 (18.3↑)} \\ \hline

\Xhline{1.2pt}
\end{tabular}}
\end{table}

\begin{table}[htbp]
\small
\centering
\caption{The two-stage correction strategy of our UFEM.}
\renewcommand\arraystretch{1.1}
\label{tab:twostage2}
\scalebox{0.83}{
\begin{tabular}{l||cccc}
\Xhline{1.2pt}
                    & \multicolumn{4}{c}{\textbf{\emph{UFEM\_Two-stage (\%)}}} \\
\multirow{-2}{*}{\textbf{Datasets}}     & \textbf{Baseline}       & \textbf{S1}         & \textbf{S2}     & \cellcolor[HTML]{EAEAEA}\textbf{S1+S2}   \\ \hline
\textbf{ExDARK (R)}        & 46.3    & 52.0 (5.7↑)      & 53.5 (7.2↑)   & \cellcolor[HTML]{EAEAEA}\textbf{54.2 (7.9↑)}    \\
\textbf{RTTS (Y)}       & 23.7    & 28.4 (4.7↑)       & 27.8 (4.1↑)   & \cellcolor[HTML]{EAEAEA}\textbf{28.8 (5.1↑)}    \\
\Xhline{1.2pt}
\end{tabular}}
\end{table}

\begin{figure}
    \centering
    \includegraphics[width=0.96\linewidth, height=0.14\textheight]{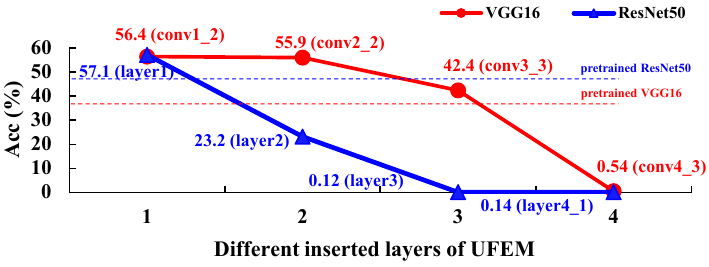}
    \caption{The ablation study of different layers of UFEM inserted into the backbone, which is mainly performed on Fog3 of ImageNet-C.}
    \label{fig:layers}
\end{figure}

\textbf{Ablation of UFEM Insertion Layers.} Further, we conduct an ablation to verify the feature correction effects achieved by integrating UFEM into various layers of the backbone. Specifically, UFEM of varying scales is respectively inserted into “conv1\_2”, “conv2\_2”, “conv3\_3”, and “conv4\_3” of VGG16, or “layer1”, “layer2”, “layer3”, and “layer4\_1” of ResNet50 to evaluate their performance on Fog3 of ImageNet-C. As reported in Fig. \ref{fig:layers}, it can be observed that inserting UFEM into the shallow layers of network tends to bring positive accuracy improvements. However, as the insertion layers deepen, the accuracy gains gradually diminish, and even falling far below the baseline. This is because that deep-layer features primarily encode semantic-related cues, posing challenges in learning the generalized mapping relationship that translates degraded features into clear ones. Therefore, inserting our UFEM into the shallow layers of network is a superior choice, which facilitates the restoration of low-level features such as edges, colors, textures, \emph{etc}, thereby boosting recognition \cite{wang2020deep}\cite{dodge2016understanding}\cite{zeiler2014visualizing}.

\textbf{Ablation of Content Consistency and Correlation Consistency.} Next, to verify the necessity of content consistency loss in Stage-2, we perform $L_{content}$ ablation. As presented in Tab. \ref{tab:content}, removing content consistency constraints and solely aligning the channel correlation matrices between EF and CF results in a sharp decline in accuracy. This decline is observed in both ResNet50 on ExDARK and YOLOv5 on RTTS, with the accuracy notably lower than Stage-1. This underscores the risk of content loss in enhanced features when only channel correlation matrices are matched, highlighting the necessity of $L_{content}$ in Stage-2. Besides, we also conduct an ablation on $L_{correlation}$, exploring three implementation forms: L1 distance, KL divergence, and Cosine similarity. As shown in Tab. \ref{tab:content}, employing KL divergence to constrain the channel correlation matrices generally leads to negative results. Instead, measuring cosine similarity of matrices brings positive improvements, although not the best ones. In contrast, minimizing the L1 distance of matrices, always delivers optimal performance, underscoring the rationale behind our UFEM.

\begin{table}[htbp]
\small
\centering
  \caption{The ablation study of the content-consistent loss and correlation-consistent loss in Stage-2. }
  \label{tab:content}
  \scalebox{0.76}{
    \begin{tabular}{l||cccc||c||c}
    \Xhline{1.5pt}
    \multirow{2}{*}{\textbf{Method}}       & \multirow{2}{*}{\textbf{L\_content}} & \multicolumn{3}{c||}{\textbf{L\_correlation}}             & \multirow{2}{*}{\textbf{Acc (\%)}} & \multirow{2}{*}{\textbf{mAP (\%)}} \\ \cline{3-5}
                                  &                             & \textbf{L1} & \textbf{KL} & \textbf{Cosine} &                                 &                                 \\ \hline
    \textbf{Baseline}                &            &             &               &                   & 46.3                            & 23.7                            \\ 
    \textbf{Stage-1}                  &                             &             &               &                   & \textbf{52.0}                            & \textbf{28.4}                            \\ \cdashline{1-7}[2pt/2pt]
    \multirow{4}{*}{\textbf{S1+S2}} &                             & \Checkmark            &               &                   & \color[HTML]{f97d1c}\textbf{48.9 (3.1↓)}                            & \color[HTML]{f97d1c}\textbf{27.9 (0.5↓)}                            \\
                                  & \cellcolor[HTML]{E8E8E8}\Checkmark                           & \cellcolor[HTML]{E8E8E8}\Checkmark           & \cellcolor[HTML]{E8E8E8}              & \cellcolor[HTML]{E8E8E8}                  & \cellcolor[HTML]{E8E8E8}\color[HTML]{fe0000}\textbf{54.2 (2.2↑)}                            & \cellcolor[HTML]{E8E8E8}\color[HTML]{fe0000}\textbf{28.8 (0.4↑)}                            \\ \cdashline{2-7}[2pt/2pt]
                                  & \Checkmark                           &             & \Checkmark             &                   & 13.7                            & 27.0                            \\
                                  & \Checkmark                            &             &               & \Checkmark                  & 53.5                            & 28.7                            \\ \Xhline{1.5pt}
\end{tabular}
  }
\end{table}

\begin{figure}[tbp]
    \centering
    \includegraphics[width=0.94\linewidth, height=0.13\textheight]{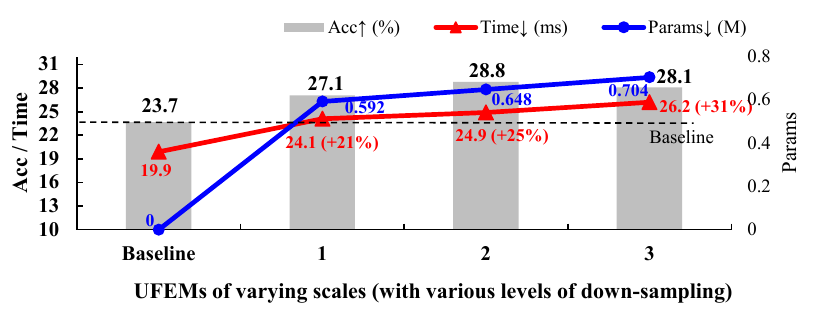}
    \caption{The computational overhead of UFEM at various scales, and the scale is determined by its down-sampling levels: 1-layer, 2-layers, and 3-layers.}
    \label{fig:time}
\end{figure}

\textbf{Computational overhead of UFEM at various scales.} Then, to explore the additional overhead posed by our UFEM, we conduct an ablation study on the computational overhead of UFEM at various scales, considering both inference time and module parameters. Specifically, we devise three variants of UFEM in U-Net architecture, incorporating 1, 2, and 3 down-sampling layers respectively, and these variants are then integrated into the existing YOLOv5 for evaluation. As shown in Fig. \ref{fig:time}, for Top-1 accuracy, it exhibits an initial rise followed by saturation as UFEM complexity increases, sometimes even showing a slight decline. For module parameters, it also increases with the UFEM complexity, adding approximately 0.6M (+8.5\%) to 0.7M (+9.9\%) parameters to YOLOv5 (7.1M). For inference time, the insertion of more complex UFEM indeed incurs additional inference time, resulting in an approximate 20\% to 30\% increase compared to original YOLOv5. One potential reason is that, to ensure the generality of our feature enhancement module, we opt to conduct feature correction at shallow layers, which typically exhibit higher spatial resolution, thereby inevitably introducing extra overhead. Therefore, for the trade-off between accuracy improvements and computational overhead, we choose a UFEM structure with 2 down-sampling layers in all experiments.

\textbf{Robustness of UFEM to varying severity.} Next, to verify the robustness of our method, we conduct an ablation that evaluates our UFEM under various degradation levels. However, due to the lack of real datasets with varied degradation levels and precise labeling, referring to \cite{belaroussi2014impact}, we manually divided RTTS into two subsets: Light\_RTTS and Dense\_RTTS, consisting of 2,705 and 1,614 images respectively. After that, we use the weights trained on whole RTTS to evaluate the robustness of UFEM on Light\_RTTS and Dense\_RTTS respectively. As shown in Tab. \ref{tab:haze}, UFEM achieves corresponding accuracy improvements in both light and dense fog, demonstrating the generality of our method. Moreover, our UFEM narrowed the performance gap of YOLOv5 on light fog and dense fog, achieving the gap from 8.9\% to 4.5\%. However, UFEM indeed exhibits lower accuracy gains in light fog compared to dense fog, indicating a need for further improvement of our UFEM in handling mild degradation.

\begin{figure}[htbp]
  \begin{minipage}[b]{0.5\textwidth}
    \centering
    \includegraphics[width=0.9\linewidth,height=0.125\textheight]{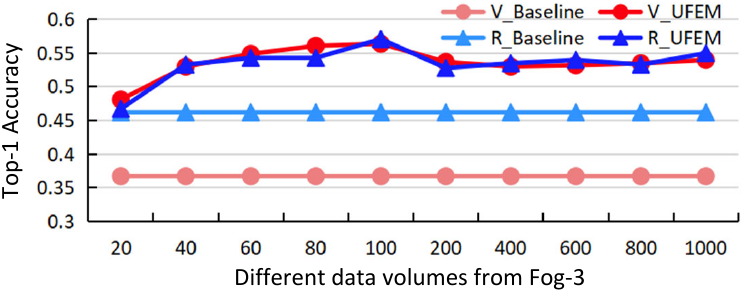}
  \end{minipage}
  \\[0.3cm]
  \begin{minipage}[b]{0.5\textwidth}
    \centering
    \includegraphics[width=0.92\linewidth,height=0.13\textheight]{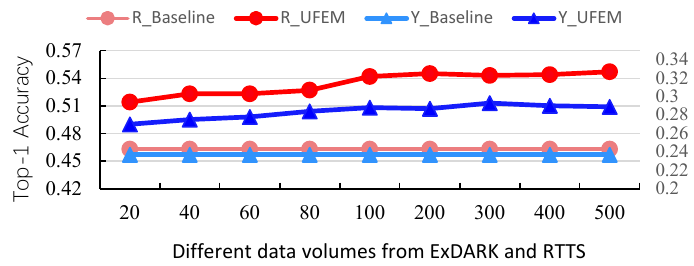}
  \end{minipage}
  \caption{The accuracy of UFEM under different data volumes with two models on both synthetic and real datasets, including VGG16 and ResNet50 on Fog3, ResNet50 on ExDARK, and YOLOv5 on RTTS.}
  \label{fig:dataAmount}
\end{figure}

\textbf{Insensitivity of Training Set Scale.} Last but not least, considering the challenge of collecting large-scale degraded images in the real world, we conduct the data sensitivity experiments by training the UFEM with different numbers of images, as shown in Fig. \ref{fig:dataAmount}. It can be observed that our method exhibits significant improvements across various data sizes, even though only a few dozen unpaired images are utilized for training, both on synthetic and real datasets. In particular, even with just 20 unpaired images for training, UFEM improves the accuracy by 5.1\% and 3.3\% on ExDARK and RTTS, respectively, demonstrating the practicality of our method in real-world degradations. Besides, as the number of images exceeds 100, the performance gains tend to saturate. Therefore, in this paper, we use 100 pairs of unpaired images to train our UFEM by default, which is a condition easily met in real-world autonomous driving. 

\begin{table}[htbp]
\small
\centering
  \caption{Performance comparison of our UFEM at different degradation levels.}
  \label{tab:haze}
  \renewcommand\arraystretch{1.25}
  \scalebox{0.8}{
   \begin{tabular}{l||ccc}
    \Xhline{1.5pt}
    \multirow{2}{*}{Method} & \multicolumn{3}{c}{\textbf{mAP on RTTS (\%)}}                         \\ \cline{2-4} 
                      & \textbf{RTTS}        & \multicolumn{1}{c}{\textbf{Light\_RTTS}} & \textbf{Dense\_RTTS} \\ \hline
    Baseline          & 23.7        & 26.8                             & 17.9        \\ \hline
    YOLOv5 UFEM            & \textbf{28.8 (+5.1↑)} & \color[HTML]{3531ff} \textbf{30.5 (+3.7↑)}                      & \color[HTML]{fe0000}\textbf{26.0 (+8.1↑)} \\ \Xhline{1.5pt}
    \end{tabular}
  }
\end{table}

\subsection{Further Expansion}
\textbf{Generalization of UFEM in Vision Transformer.} To further explore the applicability of our method, we extend UFEM to Vision Transformer architectures. Actually, our DCP, which is a prior based on convolutional features, renders it unsuitable for a pure Transformer. However, we have verified the effectiveness of integrating our UFEM into hybrid CNN and Transformer architectures. Specifically, we adopt the EdgeNeXt \cite{maaz2022edgenext} pre-trained on ImageNet as our baseline, and integrate our UFEM into its shallow layers to evaluate  the effectiveness of our method on two degradation types: fog and motion blur. As shown in Tab. \ref{tab:trans}, applying the EdgeNeXt pretrained on clear images to degraded ones still results in inevitable accuracy degradation. In contrast, integrating our UFEM into existing EdgeNeXt significantly improves its performance on various degradation levels, such as an 18.8\% improvement on fog5 and a 5.0\% improvement on fog1. These results indicate that our UFEM applies to both CNN and hybrid CNN-Transformer architectures, but it is indeed unsuitable for pure Transformer, representing a limitation.

\begin{figure}[htbp]
\centering
\includegraphics[width=0.96\linewidth]{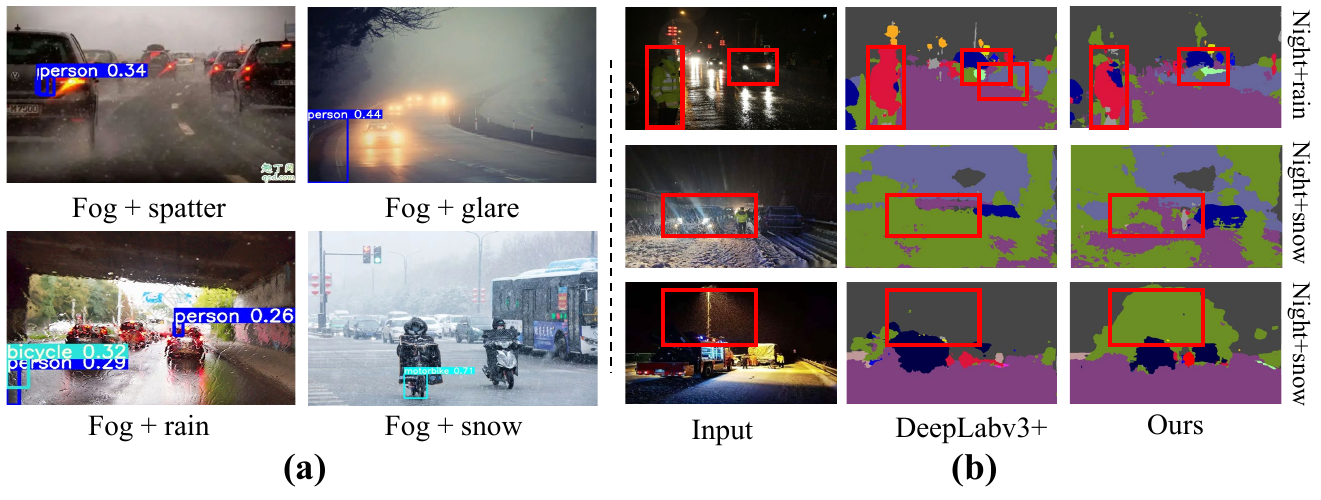}
\caption{Some failure cases of our method. The UFEM is less effective for mixed degradation scenarios.}
\label{fig:failed}
\end{figure}

\textbf{Some Failure Cases for Further Improvement.} To further explore the potential areas of our method for future improvement, we present some failure cases where our UFEM may perform poorly. As shown in Fig. \ref{fig:failed}(a), since our UFEM focuses on rectifying a specific type of degraded features into clear ones, it is susceptible to various mixed degradation cues, such as fog and spatter, fog and snow, \emph{etc}, resulting in numerous false or missed detections. As shown in Fig. \ref{fig:failed}(b), our UFEM also suffers from mixed degradation scenarios in semantic segmentation, such as night and rain, night and snow, typically resulting in pixel misclassifications due to ineffective correction of mixed degradation features. Overall, our UFEM is less effective for mixed degradation scenarios, emphasizing the focus of our future endeavors.

  


\begin{table}[htbp]
\normalsize
\centering
  \caption{Integrating our UFEM into existing CNN-Transformer network and evaluating on two different degradation types.}
  \label{tab:trans}
  \renewcommand\arraystretch{1.05}
   \scalebox{0.82}{
   \begin{tabular}{l||ccc}
    \Xhline{1.5pt}
    \multirow{2}{*}{\textbf{Method}} & \multicolumn{3}{c}{\textbf{Top-1 Acc on Fog (\%)}}          \\ \cline{2-4} 
                            & Level-1        & Level-3        & Level-5         \\ \hline
    Baseline                & 66.3            & 53.8           & 31.2            \\ \hline
    E UFEM                  & \textbf{71.3 (+5.0)}     & \textbf{60.6 (+6.8)}    & \textbf{50.0 (+18.8)}    \\ \hline \hline
    \multirow{2}{*}{\textbf{Method}} & \multicolumn{3}{c}{\textbf{Top-1 Acc on Motion\_blur (\%)}} \\ \cline{2-4} 
                            & Level-1         & Level-3        & Level-5         \\ \hline
    Baseline                & 70.0            & 50.0           & 24.9            \\ \hline
    E UFEM                  & \textbf{74.6 (+4.6)}     & \textbf{59.5 (+9.5)}    & \textbf{35.9 (+11.0)}    \\ \Xhline{1.5pt}
    \end{tabular}}
\end{table}

\section{Conclusion}
This paper focuses on addressing the decline in environmental perception performance posed by real-world degradations, which could seriously affect the visibility and perceptual capabilities of visual recognition system in autonomous vehicles. To improve the performance of existing models on degraded images, we propose the Deep Channel Prior (DCP) for degraded visual recognition, which is based on the observation that the feature channel correlations exhibit distinct margins according to degradation types, regardless of the image content and semantics. Based on this, a plug-and-play Unsupervised Feature Enhancement Module (UFEM) is devised to restore latent content, suppress artifacts, and modulate channel correlation under the guidance of DCP. The UFEM demonstrates efficacy even when trained on a modest dataset, \emph{e.g.}, 100 unpaired clear and degraded images. Experiments on three vision tasks and eight benchmark datasets demonstrate that UFEM can effectively improve the performance of existing models under various real-world degradations.



\vspace{11pt}

\bibliographystyle{IEEEtran}
\bibliography{main} 

\begin{IEEEbiography}
[{\includegraphics[width=1in,height=1.25in,clip,keepaspectratio]{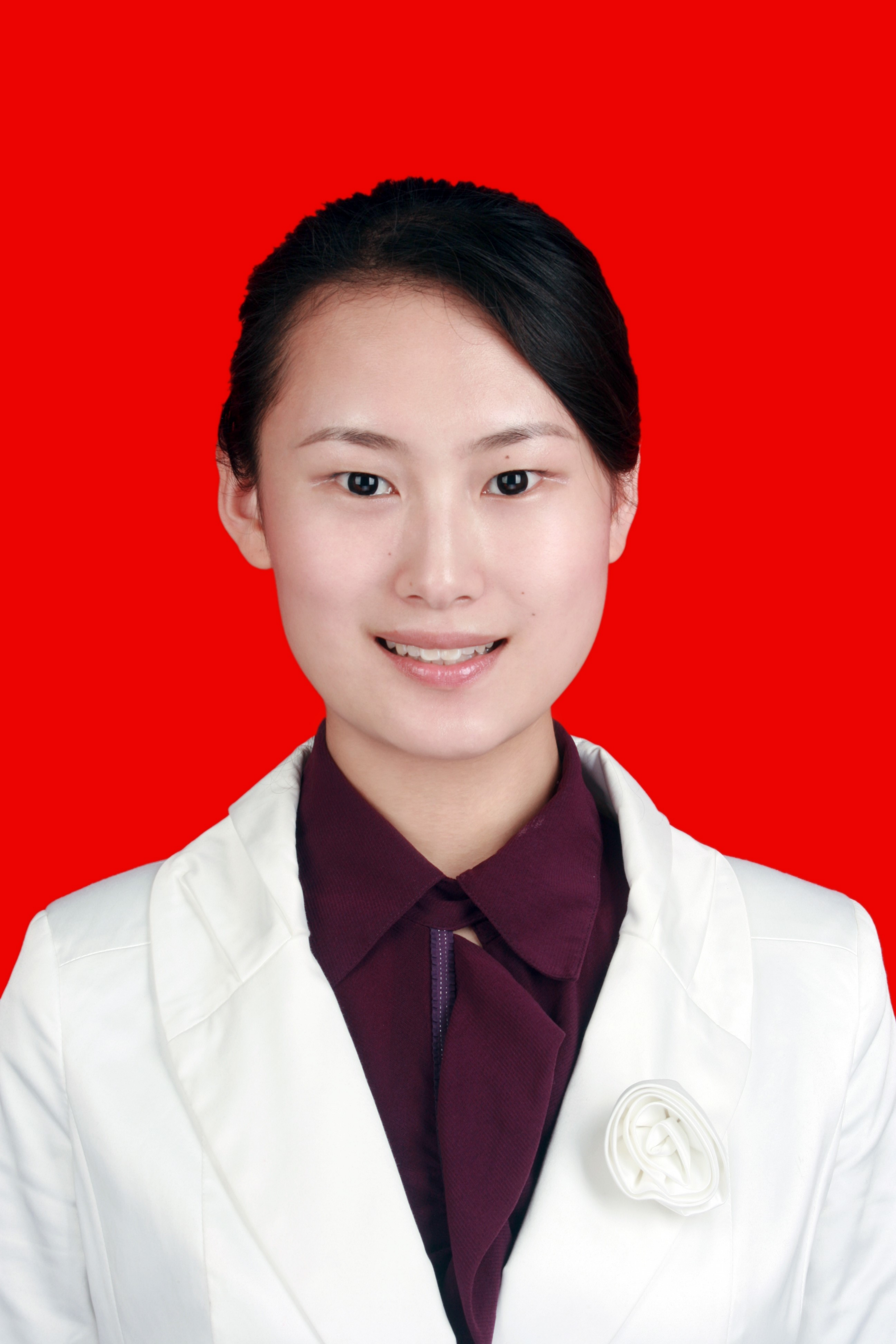}}] 
{Zhanwen Liu} (Member, IEEE) received the B.S. degree from Northwestern Polytechnical University, Xi'an, China, in 2006, the M.S. and the Ph.D. degrees in Traffic Information Engineering and Control from Chang'an University, Xi'an, China, in 2009 and 2014 respectively. She is currently a professor with School of Information Engineering, Chang'an University, Xi'an, China. Her research interests include vision perception, autonomous vehicles, deep learning and intelligent transportation systems.
\end{IEEEbiography}

\begin{IEEEbiography}
[{\includegraphics[width=1in,height=1.25in,clip,keepaspectratio]{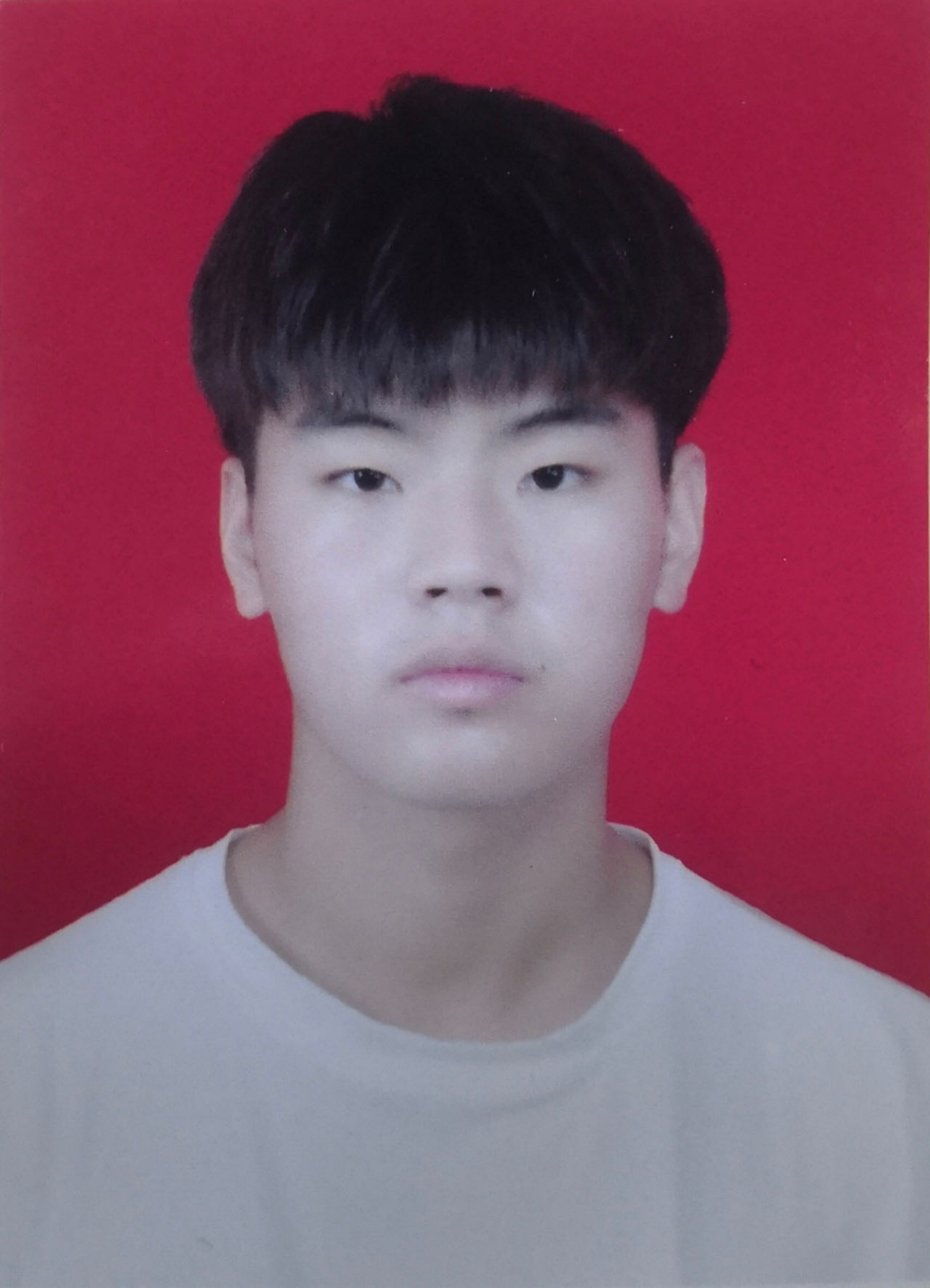}}] 
{Yuhang Li} received the B.S. degree from Chang’an University in Xi’an, China, in 2022, where he is currently working toward the M.S. degree with the Department of Computer Science and Technology. His current research interests include domain adaptation, object detection under adverse weather conditions, and their applications in intelligent vehicle and road infrastructure perception.
\end{IEEEbiography}

\begin{IEEEbiography}
[{\includegraphics[width=1in,height=1.25in,clip,keepaspectratio]{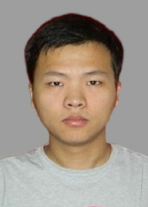}}] 
{Yang Wang} (Member, IEEE) received the Ph.D. degree in control science and engineering from University of Science and Technology of China, in 2021. He is currently an associate professor with the School of Information Engineering, Chang'an University, Xi'an, China. His research interests include machine learning and image processing.
\end{IEEEbiography}

\begin{IEEEbiography}
[{\includegraphics[width=1in,height=1.25in,clip,keepaspectratio]{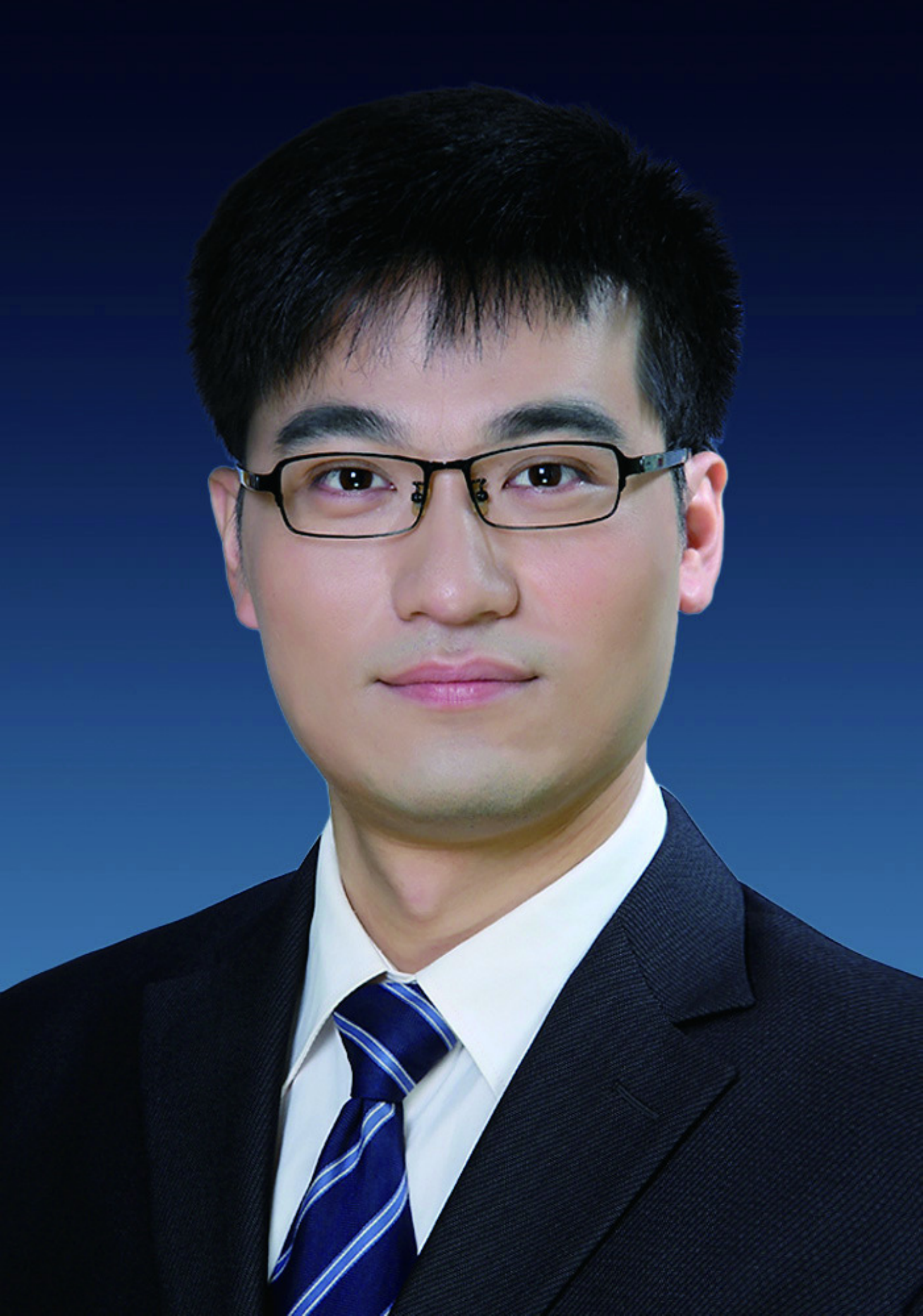}}] 
{Bolin Gao} received the B.S. and M.S. degrees in vehicle engineering from Jilin University, Changchun, China, in 2007 and 2009, respectively, and the Ph.D. degree in vehicle engineering from Tongji University, Shanghai, China, in 2013. He is currently an Associate Research Professor with the School of Vehicle and Mobility, Tsinghua University, Beijing, China. His research interests include the theoretical research and engineering application of the dynamic design and control of Intelligent and connected vehicles, Especially about collaborative perception and tracking method in cloud control system, intelligent predictive cruise control system on commercial trucks with cloud control mode, and the test and evaluation of intelligent vehicle driving system.
\end{IEEEbiography}

\begin{IEEEbiography}
[{\includegraphics[width=1in,height=1.25in,clip,keepaspectratio]{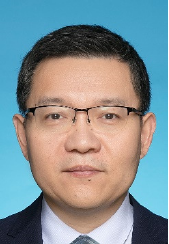}}] 
{Yisheng An} received the M.S. and Ph.D. degrees in systems engineering from Xi’an Jiaotong University, Xi’an, China, in 2001 and 2007, respectively. He is an IEEE Member, and a Professor with the Department of Computer Science and Engineering, School of Information Engineering, Chang’an University, Xi’an. His research interests include Petri nets, internet of Vehicles, intelligent transportation systems and distributed information systems.
\end{IEEEbiography}

\begin{IEEEbiography}
[{\includegraphics[width=1in,height=1.25in,clip,keepaspectratio]{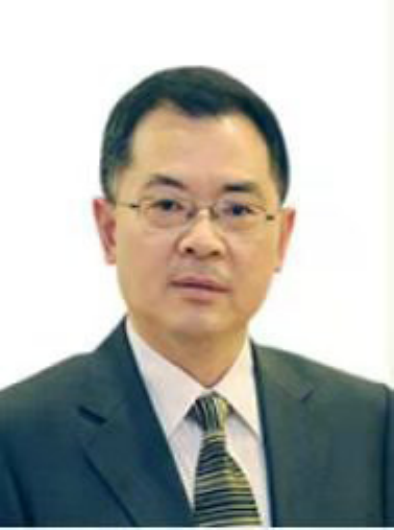}}] 
{Xiangmo Zhao} (Member, IEEE) received the B.S degree from Chongqing University, China, in 1987, and the M.S. and Ph.D. degrees from Chang' an University, China, in 2002 and 2005, respectively. He is currently a distinguished professor with School of Information Engineering, Chang'an University, Xi'an, China. His research interests include intelligent transportation systems, internet of vehicles, connected and autonomous vehicles testing technology.
\end{IEEEbiography}

\vfill

\end{document}